\documentclass[10pt,twocolumn,letterpaper]{article}

\usepackage{iccv}
\usepackage{times}
\usepackage{epsfig}
\usepackage{graphicx}
\usepackage{amsmath}
\usepackage{amssymb}
\usepackage{amsfonts}
\usepackage{subcaption}
\usepackage{tabularx}
\usepackage{bbm}

\usepackage{mathtools}
\usepackage{multirow}
\makeatletter

\graphicspath{{images/}}
\def\BState{\State\hskip-\ALG@thistlm}

\newcommand{\comment}[1]{}
\newcommand{\parag}[1]{\vspace{-3mm}\paragraph{#1}}

\newcommand{\mB}[0]{\mathcal{B}}
\newcommand{\mE}[0]{\mathcal{E}}
\newcommand{\mL}[0]{\mathcal{L}}
\newcommand{\mT}[0]{\mathcal{T}}
\newcommand{\mM}[0]{\mathcal{M}}
\newcommand{\mJ}[0]{\mathcal{J}}
\newcommand{\mG}[0]{\mathcal{G}}
\newcommand{\mW}[0]{\mathcal{W}}

\newcommand{\bB}[0]{\mathbf{B}}
\newcommand{\bF}[0]{\mathbf{F}}
\newcommand{\bG}[0]{\mathbf{G}}
\newcommand{\bM}[0]{\mathbf{M}}
\newcommand{\bN}[0]{\mathbf{N}}

\newcommand{\bOne}[0]{\mathbbm{1}}

\newcommand{\Garnet}[0]{{\bf GarNet}}
\newcommand{\Global}[0]{{\bf GarNet-Global}}
\newcommand{\Local}[0]{{\bf GarNet-Local}}
\newcommand{\Late}[0]{{\bf GarNet-Naive}}

\newcommand{\PBS}[0]{{\bf PBS}}

\newif\ifdraft
\drafttrue

\newcommand{\MS}[1]{\ifdraft {\color{green}{\textbf{MS: #1}}}\else {}\fi}

\usepackage[pagebackref=true,breaklinks=true,letterpaper=true,colorlinks,bookmarks=false]{hyperref}
\usepackage{flushend}
\iccvfinalcopy 

\def\iccvPaperID{701} 

\ificcvfinal\fi
\begin{document}

\title{GarNet: A Two-Stream Network for Fast and Accurate 3D Cloth Draping}

\author{Erhan Gundogdu$^1$, Victor Constantin$^1$, Amrollah Seifoddini$^2$\\
\and
Minh Dang$^2$, Mathieu Salzmann$^1$, Pascal Fua$^1$ \vspace{0.2cm}\\
$^1$CVLab, EPFL, Switzerland\\
$^2$Fision Technologies, Zurich, Switzerland\\
{\tt\small \{erhan.gundogdu, victor.constantin, mathieu.salzmann, pascal.fua\}@epfl.ch}\\
{\tt\small \{amrollah.seifoddini, minh.dang\}@fision-technologies.com}
}

\maketitle
\ificcvfinal\thispagestyle{empty}\fi
\begin{abstract}
While Physics-Based Simulation (PBS) can accurately drape a 3D garment on a 3D body, it remains too costly for real-time applications, such as virtual try-on. By contrast, inference in a deep network, requiring a single forward pass, is much faster. Taking advantage of this, we propose a novel architecture to fit a 3D garment template to a 3D body. Specifically, we build upon the recent progress in 3D point cloud processing with deep networks to extract garment features at varying levels of detail, including point-wise, patch-wise and global features. We fuse these features with those extracted in parallel from the 3D body, so as to model the cloth-body interactions. The resulting two-stream architecture, which we call as GarNet, is trained using a loss function inspired by physics-based modeling, and delivers visually plausible garment shapes whose 3D points are, on average, less than 1 cm away from those of a PBS method, while running 100 times faster. Moreover, the proposed method can model various garment types with different cutting patterns when parameters of those patterns are given as input to the network.

\comment{
	Virtual try-on has become popular with the recent progress of virtual reality and online shopping. Realistic cloth simulation is required by these applications where Physics-Based Simulations (PBS) can provide plausible solutions but with a considerable computational cost. To accelerate the computation, there exist data-driven methods most of which require pre- and post-processing steps. To bypass these limitations, we propose a deep learning method that reproduces the output of an existing PBS with a two-stream architecture taking into account both the template garment mesh and the current body shape. Our novel architecture considers a hierarchy of features by using the recent advancements of deep learning for 3D point clouds and meshes. Our method directly produces the final fitted garment by incorporating the constrains of PBS approaches at training time which makes it aware of physical constraints such as garment-body interpenetration.  To train and test the proposed approach, we have curated a garment simulation dataset consisting of different body shapes and poses on four different garment types. Our extensive experiments on our dataset and a recently published one demonstrate that the proposed method can predict the fitted garments with less than 1.5 cm. of vertex distance and 12 degrees of face normal error while maintaining the realism.
}
\end{abstract}


\section{Introduction}

\begin{figure}[ht!]
	\centering
	\includegraphics[width=0.75\linewidth]{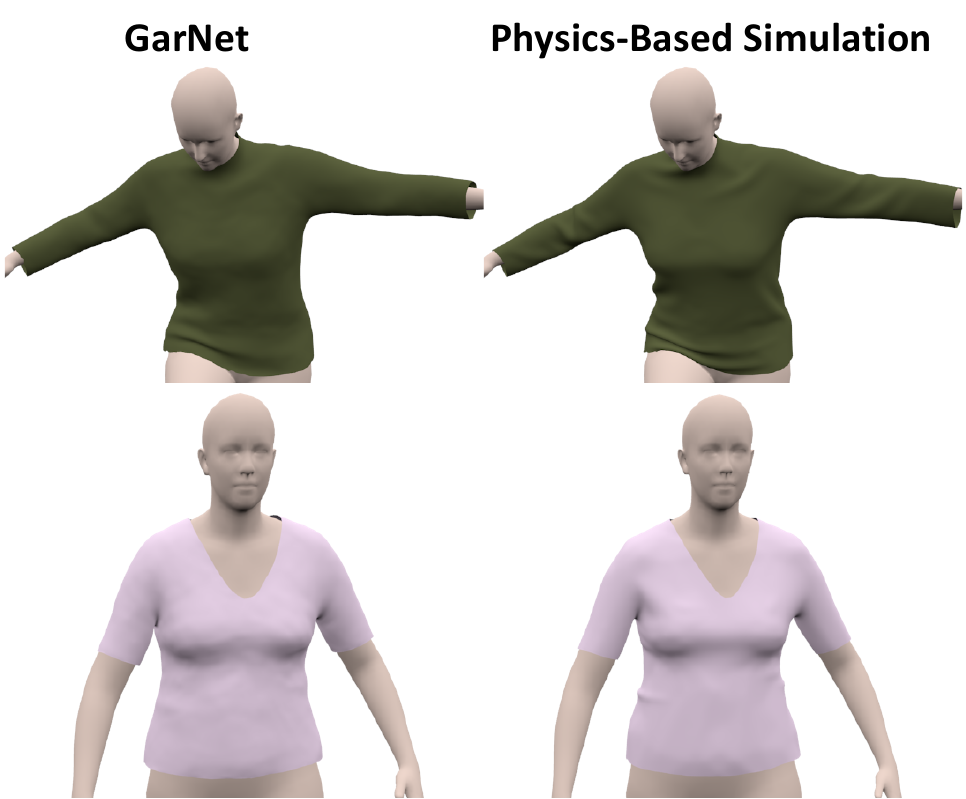}
	\vspace{-3mm}
	\caption{\small {Draping a sweater and a T-shirt. Our method produces results as plausible as those of a PBS method, but runs 100x faster.}}
	\label{fig:teaser}
\end{figure}

Garment simulation is useful for many purposes such as virtual try-on, online shopping, gaming, and virtual reality. Physics-Based Simulation (PBS) can deliver highly realistic results, but at the cost of heavy computation, which makes it unsuitable for real-time and web-based applications. As shown in Fig.~\ref{fig:teaser}, in this paper, we propose to train a deep network to produce visually plausible 3D draping results, as achieved by PBS, but much faster. {\let\thefootnote\relax\footnotetext{This work was supported in part by the CTI Project 26455.1 PFES-ES.}}\par


\begin{figure*}[htbp]
	\centering
	\begin{subfigure}[b]{0.42\textwidth}
		\includegraphics[width=\textwidth]{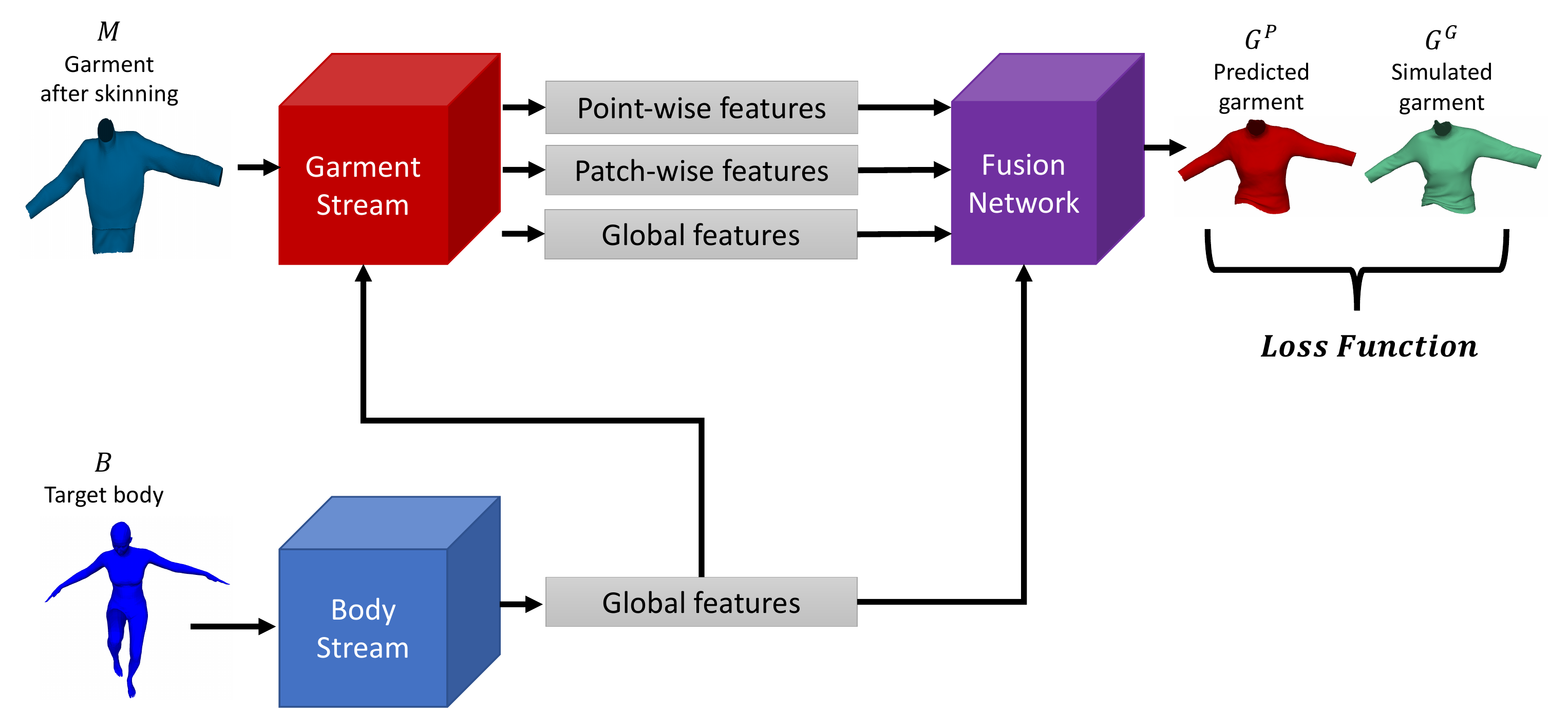}
		\caption{\small \Global}
		\label{fig:streamsA}
	\end{subfigure}
	\begin{subfigure}[b]{0.42\textwidth}
		\includegraphics[width=\textwidth]{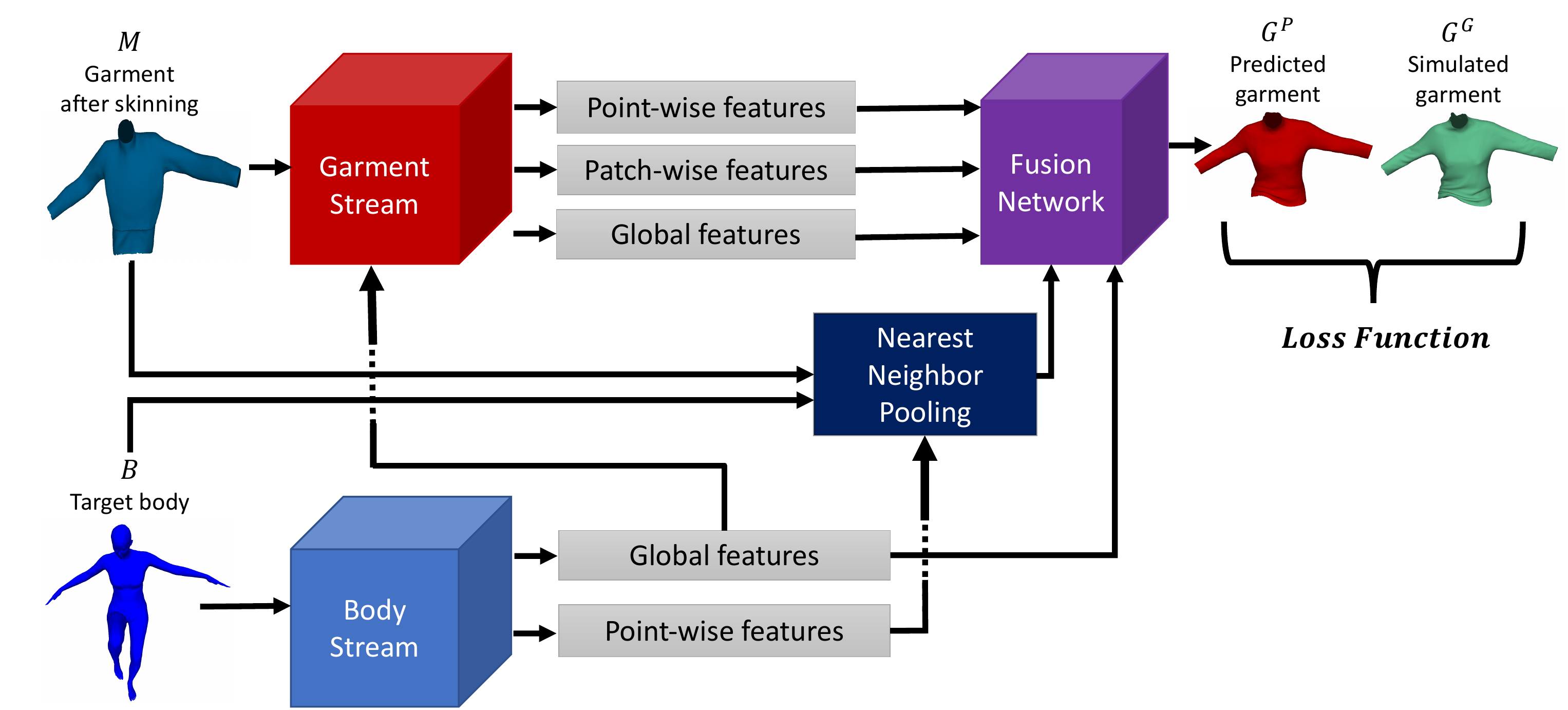}
		\caption{\small \Local}
		\label{fig:streamsB}
	\end{subfigure}
	\vspace{-3mm}
	\caption{ {\bf Two versions of our \textbf{GarNet}.} Both take as input a target body and the garment mesh roughly aligned with the body pose by using~\cite{Kavan07}. \Global{}: We fuse the global body features with the garment ones both early and late. \Local{}: In addition, we use a nearest neighbor pooling for local body features and feed the result to the fusion network to combine the body and garment features.}
	\label{fig:streams}
\end{figure*}

Realistic simulation of cloth draping over the human body requires accounting for the global 3D pose of the person and for the local interactions between skin and cloth caused by the body shape. To this end, we introduce the architecture depicted by Fig.~\ref{fig:streams}. It consists of a garment stream and a body stream. The body stream uses a PointNet~\cite{Qi17a} inspired architecture to extract local and global information about the 3D body. The garment stream exploits the global body features to compute point-wise, patch-wise and global features for the garment mesh. These features, along with the global ones obtained from the body, are then fed to a fusion subnetwork to predict the shape of the fitted garment. In one implementation of our approach, shown in  Fig.~\ref{fig:streamsA}, the local body features are only used {\it implicitly} to compute the global ones. In a more sophisticated implementation, we {\it explicitly} take them into account to further model the skin-cloth interactions. To this end, we introduce an auxiliary stream that first computes the $K$ nearest body vertices for each garment vertex, performs feature pooling on point-wise body features and finally feeds them to the fusion subnetwork. This process is depicted by Fig.~\ref{fig:streamsB}. We will see that it performs better than the simpler one, indicating that local feature pooling is valuable.



By incorporating appropriate loss terms in the objective function that we minimize during training, at test time, we avoid the need for extra post-processing steps to minimize cloth-body interpenetration and undue tightness that PBS tools~\cite{Nvcloth,Optitext,NvFlex,MarvelousDesigner}, optimization-based~\cite{Brouet12} and data-driven~\cite{Guan12,Wang18} methods often require. Furthermore, by relying on convolution and pooling operations, our approach naturally scales to point clouds of arbitrary resolution. This is in contrast to data-driven methods~\cite{Guan12,Wang18} that rely on a low-dimensional subspace whose size would typically need to grow as the resolution increases, thus strongly affecting these models' memory requirements.

Our contribution is therefore a novel architecture for static garment simulation that delivers fitting results in real-time by properly modeling the body and garment interaction, thus reducing cloth and body interpenetration. For training purposes, we built a dataset that will be made public\footnote{Please check for the dataset at \url{https://cvlab.epfl.ch/research/garment-simulation/garnet/}}. It comprises a pair of jeans, a t-shirt and a sweater worn by 600 bodies from the SMPL dataset~\cite{Loper15} in various poses. Experiments on our dataset show that our network can effectively handle many body poses and shapes. Moreover, our approach can incorporate additional information, such as cutting patterns, when available. To illustrate this, we make use of the recently-published data of~\cite{Wang18}, which contains different garment types with varying cutting patterns. Our experiments demonstrate that our method outperforms the state-of-the-art one of~\cite{Wang18} on this dataset.
Finally, whereas the PBS approach that we take as reference takes more than 10 seconds to predict the shape of a garment, ours takes less than 70 ms, thus being practical for real-time applications.


\comment{ 

}


\section{Related Work}
\label{sec:related}

Many professional tools can model cloth deformations realistically using Physics-Based Simulation (PBS)~\cite{Nvcloth,Optitext,NvFlex,MarvelousDesigner}. However, they are computationally expensive, which precludes real-time use. Furthermore, manual parameter tuning is often required. First, we briefly review recent approaches to overcoming these limitations. Then, we summarize the deep network architectures for 3D point cloud and mesh processing, and the related works for 3D human/cloth modeling.

\parag{Data-Driven Approaches.}
They are computationally less intensive and memory demanding, at least at run-time, and have emerged as viable competitors to PBS. One of the early methods~\cite{Kim08} relies on generating a set of garment-body pairs. At test time, the garment shape in an unseen pose is predicted by linearly interpolating the garments in the database. An earlier work \cite{Miguel12} proposes a data-driven estimation of the physical parameters of the cloth material while \cite{Kim13a} constructs a finite motion graph for detailed cloth effects. In~\cite{Wang10f}, potential wrinkles for each body joint are stored in a database so as to model fine details in various body poses. However, it requires performing this operation for each body-garment pair. To speed up the computation, the cloth simulation is modeled in a low-dimensional linear subspace as a function of 3D body shape, pose and motion in \cite{Aguiar10}. \cite{Guan10c} also models the relation between 2D cloth deformations and corresponding bodies in a low-dimensional space. \cite{Guan12} extends this idea to 3D shapes by factorizing the cloth deformations according to what causes them, which is mostly shape and pose. The factorized model is trained to predict the garment's final shape. \cite{Santesteban19} trains an MLP and an RNN to model the cloth deformations by decomposing them as static and dynamic wrinkles. Both~\cite{Guan12} and \cite{Santesteban19}, however,  require an {\it a posteriori} refinement to prevent cloth-body interpenetration. In a recent approach,~\cite{Wang18} relies on a deep encoder-decoder model to create a joint representation for bodies, garment sewing patterns, 2D sketches and garment shapes. This defines a mapping between any pair of such entities, for example body-garment shape. However, it relies on a Principal Component Analysis (PCA) representation of the garment shape, thus reducing the accuracy. In contrast to~\cite{Wang18}, our method operates directly on the body and garment meshes, removing the need for such a limiting representation. We will show that our predictions are more accurate as a result.

Cloth fitting has been performed using 4D data scans as in \cite{Lahner18,Pons-Moll17}. In \cite{Pons-Moll17}, garments deforming over time are reconstructed using 4D data scans and the reconstructions are then retargeted to other bodies without accounting for physics-based clothing dynamics. Unlike in \cite{Pons-Moll17}, we aim not only to obtain visually plausible results but also to emulate PBS for cloth fitting.  In \cite{Lahner18}, fine wrinkles are generated by a conditional Generative Adversarial Network (GAN) that takes as input predicted, low-resolution normal maps. This method, however, requires a computationally demanding step to register the template cloth to the captured 4D scan, while ours needs only to perform skinning of the template garment shape using the efficient method of~\cite{Kavan07}.

\parag{Point Cloud and Mesh Processing.}
A key innovation that has made our approach practical is the recent emergence of deep architectures that allow for the processing of point clouds~\cite{Qi17a,Qi17b} and meshes~\cite{Verma18}. PointNet~\cite{Qi17a,Qi17b} was the first to efficiently represent and use unordered point clouds for 3D object classification and segmentation. It has spawned several approaches to point-cloud upsampling~\cite{Yu18a}, unsupervised representation learning~\cite{Yang18a}, 3D descriptor matching~\cite{Deng18}, and finding 2D correspondences~\cite{Yi18a}. In our architecture, as in PointNet, we use Multilayer Perceptrons (MLPs) for point-wise processing and max-pooling for global feature generation. However, despite its simplicity and representative power, point-wise operations in PointNet~\cite{Qi17a}
is not sufficient to produce visually plausible garment fitting results, as we experimentally demonstrate by qualitative and quantitative analysis.

Given the topology of the point clouds, for example in the form of a triangulated mesh, graph convolution methods, unlike PointNet~\cite{Qi17a}, can produce local features, such as those of~\cite{Boscaini16,Masci15,Monti17} that rely on hand-crafted patch operators. FeastNet~\cite{Verma18} generalizes this approach by learning how to dynamically associate convolutional filter weights with features at the vertices of the mesh, and demonstrates state-of-the-art performance on the 3D shape correspondence problem. Similar to~\cite{Verma18}, we also use mesh convolutions to extract patch-wise garment features that encode the neighborhood geometry. However, in contrast to the methods whose tasks are 3D shape segmentation \cite{Qi17a, Qi17b} or 3D shape correspondence~\cite{Verma18, Boscaini16,Masci15,Monti17}, we do not work with a single point cloud or mesh as input, but with two: one for the body and the other for the garment, which are combined in our two-stream architecture to account for both shapes.

\parag{3D human body/cloth reconstruction.}

3D  body shapes/cloth are modeled from RGB/RGBD cameras in \cite{Zhang17b,Yang18c,Xu18,Habermann19,Alldieck18,Alldieck19,Yu18,Yu19} while garment and surface reconstruction methods from images are addressed  in surface/wrinkle reconstruction from images \cite{Danerek17,Bednarik18,Popa2009}. Moreover, generative models reconstruct cloths in \cite{Lassner17b,Han18}.


\comment{

which can  by utilizing material specific parameters such as stiffness and bending. Although these methods are effective for realistic results, manual adjustments are required and computationally expensive optimization methods are employed. In \cite{Wang18}, they use NVIDIA Flex \cite{NvFlex} for generating simulation data where they first deform the target body to the template garment pose, then the simulation is run to deform the garment and the body to the target pose as in \cite{Wu18}. Our cloth simulation and data generation process is similar to the approach in \cite{Wang18}, and this will be explained in Section \ref{DatasetGeneration}.
}

\comment{Data-driven cloth simulation: Physically-based simulation (PBS) methods are computationally expensive and they may require specialized hardware. Hence, learning models that can predict what the PBS approaches do is relatively an efficient solution as long as the learnt model is both computationally and memory friendly. }

\comment{ exploit the spectral eigen-decomposition of graph Laplacian \cite{Defferrard16, Bruna13, Henaff15}. However, those approaches based on spectral filtering are not appropriate for the generalization on 3D shapes with different topologies. To circumvent this, local filtering methods are designed for  where  are used.
}

\section{3D Garment Fitting}

\begin{figure}[t]
\centering
	\includegraphics[width=0.30\textwidth]{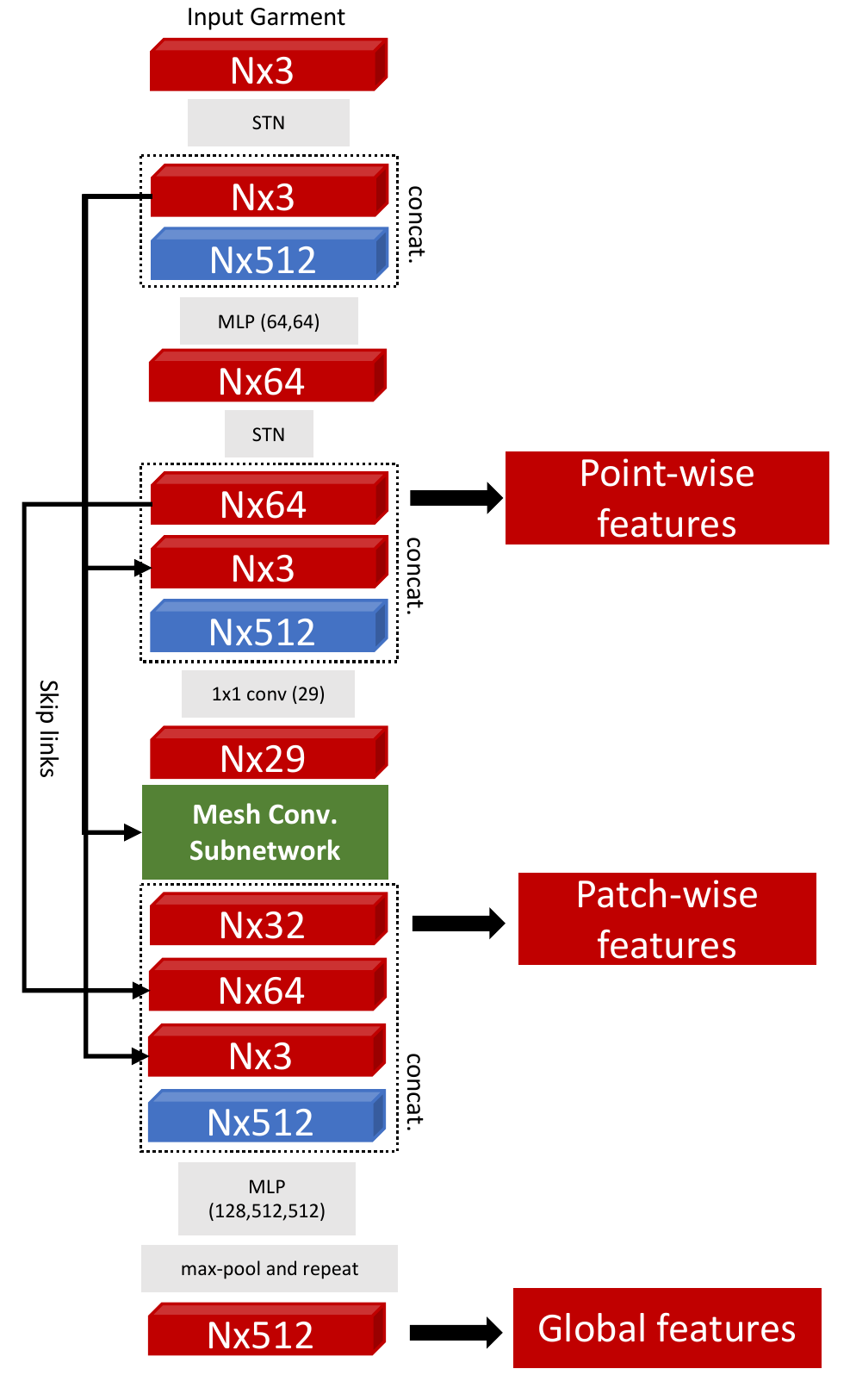}
	\vspace{-3mm}
	\caption{\small {\bf Garment branch of our network.} The grey boxes and the numbers in parenthesis denote network layers and their output channel dimensions. Red and blue ones represent garment and global body features, respectively. The green box is the mesh convolution subnetwork and depicted in more detail in Fig.~\ref{residualBlock}. STN stands for a Spatial Transformer Network used in PointNet \cite{Qi17a}. MLP blocks are shared by all $N$ points.} 
	\label{garmentBranch}
\end{figure}

To fit a garment to a body in a specific pose, we start by using a dual quaternion skinning (DQS) method~\cite{Kavan07} that produces a rough initial garment shape that depends on body pose. In this section, we introduce two variants of our \Garnet{} deep network to refine this initial shape and produce the final garment. Fig.~\ref{fig:streams} depicts these two variants.


\begin{figure}[t]
\centering
	\includegraphics[width=0.95\linewidth]{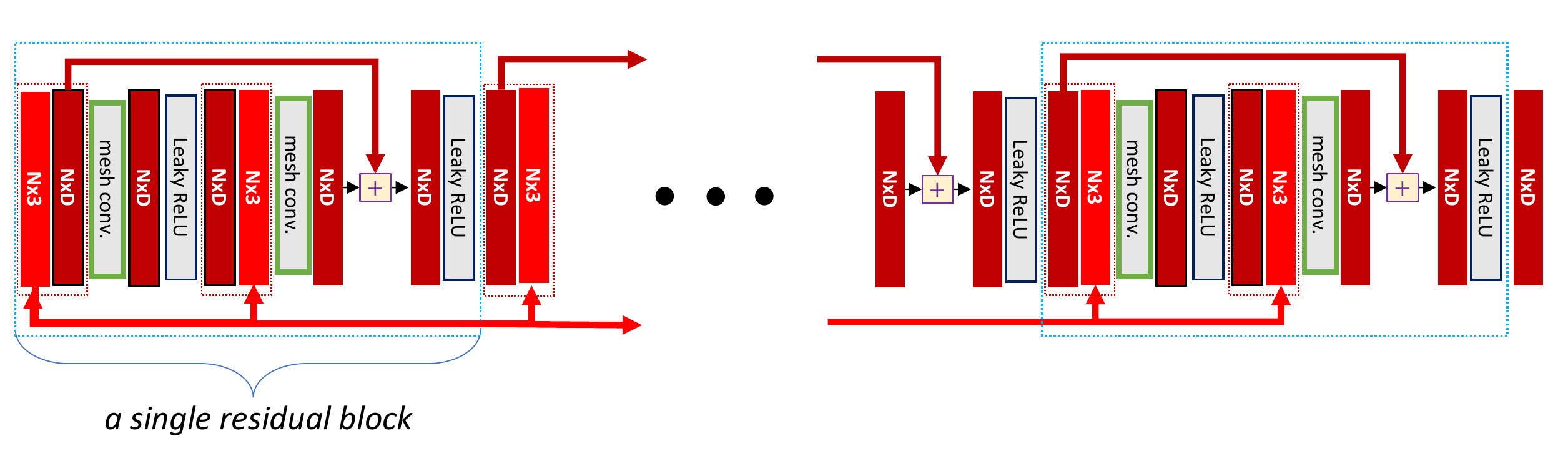}
	\vspace{-3mm}
	\caption{\textbf{Mesh conv. subnetwork.} The residual block is repeated 6 times. Dashed red rectangles indicate channel-wise concatenation. The $N\times 3$-dimensional tensors contain the 3D vertex locations of the input garment, which are passed at different stages via skip connections.} 
	\label{residualBlock}
\end{figure}

\subsection{Problem Formulation}
\label{sec:formal}

Let $\mM^0$ be the template garment mesh in the rest pose and let $\mM = \mbox{dqs}(\mM^0,\mB, \mJ_{\mM}^0, \mJ_{\mB}, \mW)$ be the garment after skinning to the body $\mB$, also modeled as a mesh, by the method~\cite{Kavan07}. Here, $\mJ_{\mM}^0$ and $\mJ_{\mB}$ are the joints of $\mM^0$ and $\mB$, respectively. $\mW$ is the skinning weight matrix for $\mM^0$. Let $f_{\theta}$ be the network with weights $\theta$ chosen so that the predicted garment $\mG^P$ given $\mM$ and $\mB$  is as close as possible to the ground-truth shape $\mG^G$. We denote the $i^{th}$ vertex of $\mM$, $\mB$, $\mG^{G}$ and $\mG^{P}$ by ${\bM}_i$, ${\bB}_i$, ${\bG_i^{G}}$ and ${\bG_i^{P}}$ $\in \mathbb{R}^3$, respectively. Finally, let $N$ be the number of vertices in $\mM$, $\mG^G$ and $\mG^P$.

Since predicting deformations from a reasonable initial shape is more convenient than predicting absolute 3D locations, we train $f_{\theta}$ to predict a translation vector for each vertex of the warped garment $\mM$ that brings it as close as possible to the corresponding ground-truth vertex. In other words, we optimize with respect to $\theta$ so that
\begin{equation}
\mT^P = f_{\theta}(\mM,\mB) \approx \mT^G \; , \label{eq:network}
\end{equation}
where $\mT^P$ and $\mT^G$ correspond to translation vectors from the skinned garment $\mM$ to the predicted and ground-truth mesh, respectively, that is $\bG_i^{P}-\bM_i$ and  $\bG_i^{G}-\bM_i$.
Therefore, the final shape of the garment mesh is obtained by adding the translation vectors predicted by the network to the vertex positions after skinning.

\subsection{Network Architecture}
\label{sec:arch}

We rely on a two-stream architecture to compute $f_{\theta}(\mM,\mB)$. The first stream, or \emph{body stream}, takes as input the body represented by a 3D point cloud while the second, or \emph{garment stream}, takes as input the garment represented by a triangulated 3D mesh. Their respective outputs are fed to a fusion network that relies on a set of MLP blocks to produce the predicted translations $\mT^P$ of Eq.~\ref{eq:network}. To not only produce a rough garment shape, but also predict fine details such as wrinkles and folds, we include early connections between the two streams, allowing the garment stream to account for the body shape even when processing local information. As shown in Fig.~\ref{fig:streams}, we implemented two different versions of the full architecture and discuss them in detail below. 

\parag{Body Stream.}
The first stream processes the body  $\mB$ in a manner similar to that of PointNet~\cite{Qi17a} (see Sec.~\ref{sec:implementation} for details). It efficiently produces point-wise and global features that adequately represent body pose and shape. Since there are no direct correspondences between 3D body points and 3D garment vertices, the global body features are key to incorporating such information while processing the garment. We observed no improvement by using mesh convolution layers in this stream.

\parag{Garment Stream.} 
The second stream takes as input the warped garment $\mM$ and the global body features extracted by the body stream to also compute point-wise and global features. As we will see in the results section, this suffices for a rough approximation of the garment shape but not to predict wrinkles and folds. We therefore use the garment mesh to create {\it patch-wise features}, that account for the local neighborhood of each garment vertex by using mesh convolution operations~\cite{Verma18}. In other words, instead of using a standard PointNet architecture, we use the more sophisticated one depicted by Fig.~\ref{garmentBranch} to compute point-wise, patch-wise, and global features. As shown in Fig.~\ref{garmentBranch}, the features extracted at each stage are forwarded to the later stages via skip connections. Thus, we directly exploit the low-level information while extracting higher-level representations.

\parag{Fusion Network.}
Once the features are produced by the garment and body streams, they are concatenated and given as input to the fusion network shown as a purple box in Fig.~\ref{fig:streams}. It consists of four MLP blocks shared by all the points, as done in the segmentation network of PointNet \cite{Qi17a}. The final MLP block outputs the 3D translations $\mT^P$ of Eq.~\ref{eq:network} from the warped garment shape $\mM$. 

\comment{Note that the global body features, denoted by blue boxes are concatenated in several layers to make the branch aware of the overall body shape. In theory, we could have done the same with the local body features but this would have massively increased the computational cost. Instead, we use them only in the fusion network. \MS{This statement is incorrect, and I think we should remove this paragraph.}}

\parag{Global and Local Variants.}


\begin{figure}[t]
\centering
	\centering
	\includegraphics[width=0.25\textwidth]{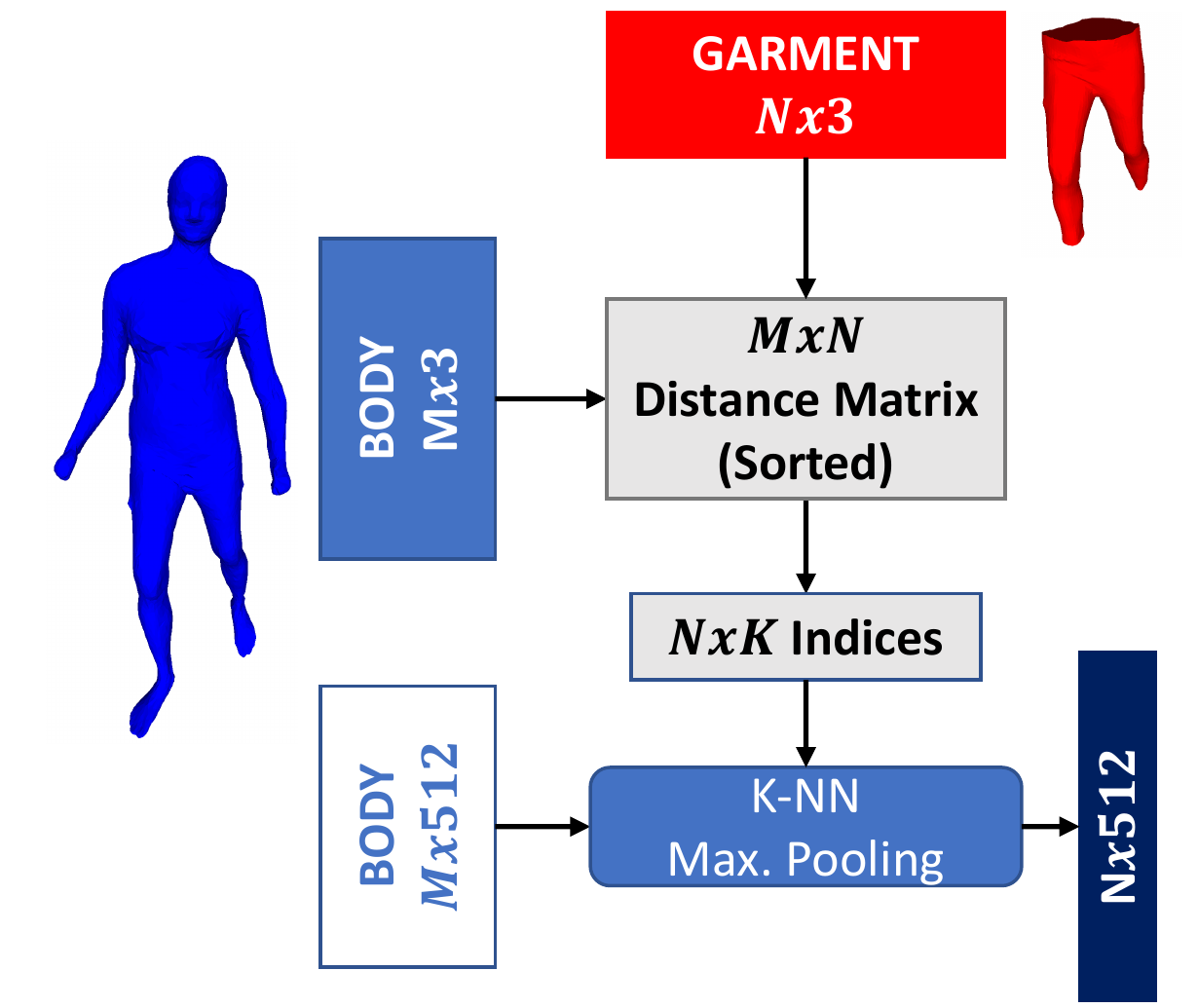}
	\vspace{-3mm}
	\caption{\small {\bf $K$ nearest neighbor pooling in Fig.~\ref{fig:streamsB}.} We compute the $K$ nearest neighbor body vertices of each garment vertex and max-pool their local features. 
	}
	\label{LocalPooling}
\end{figure}

Fig.~\ref{fig:streamsA} depicts the  \Global{} version of our architecture. It discards the point-wise body features produced by the body stream and exclusively relies on the global body ones.  Note, however, that the local body features are still implicitly used because the global ones depend on them. This enables the network to handle the garment/body dependencies without requiring explicit correspondences between body points and mesh vertices.  In the more sophisticated \Local{} architecture depicted by Fig.~\ref{fig:streamsB}, we explicitly exploit the point-wise body features by introducing a nearest neighbor pooling step to compute separate local body features for each garment vertex. It takes as input the point-wise body features and uses a nearest neighbor approach to compute additional features that capture the proximity of $\mM$ to $\mB$ and feeds them into the fusion network, along with the body and garment features. This step shown in Fig.~\ref{LocalPooling} improves the prediction accuracy due to the explicit use of local body features.

\subsection{Loss Function}
\label{sec:loss}

To learn the network weights, we minimize the loss function $\mL(\mG^{G},\mG^{P},\mB,\mM)$. We designed it to reduce the distance of the prediction  $\mG^{P}$ to the ground truth $\mG^{G}$ while also incorporating regularization terms derived from physical constraints. The latter also depend on the body $\mB$ and the garment $\mM$. We therefore write $\mL$ as
\begin{equation}
	\begin{gathered}
		L_{vertex}+\lambda_{pen}L_{pen}+\lambda_{norm}L_{norm}+\lambda_{bend}L_{bend} \; ,
	\end{gathered}
\label{eq:lossTotal}
\end{equation}
where $\lambda_{pen}$, $\lambda_{norm}$, and $\lambda_{bend}$ are weights associated with the individual terms described below. We will study the individual impact of these terms in the results section. 

\parag{Data Term.}
We take $L_{vertex}$ to be the average $L^2$ distance between the vertices of $\mG^{G}$ and $\mG^{P}$, 
\begin{equation}
		\frac{1}{N}\sum\limits_{i=1}^N {\left\lVert \bG_i^{G}-\bG_i^{P} \right\rVert}^2,
\end{equation}
where $N$ is the total number of vertices. 

\begin{figure}[t!]
	\centering
	\begin{subfigure}[b]{0.35\columnwidth}
		\includegraphics[width=\textwidth]{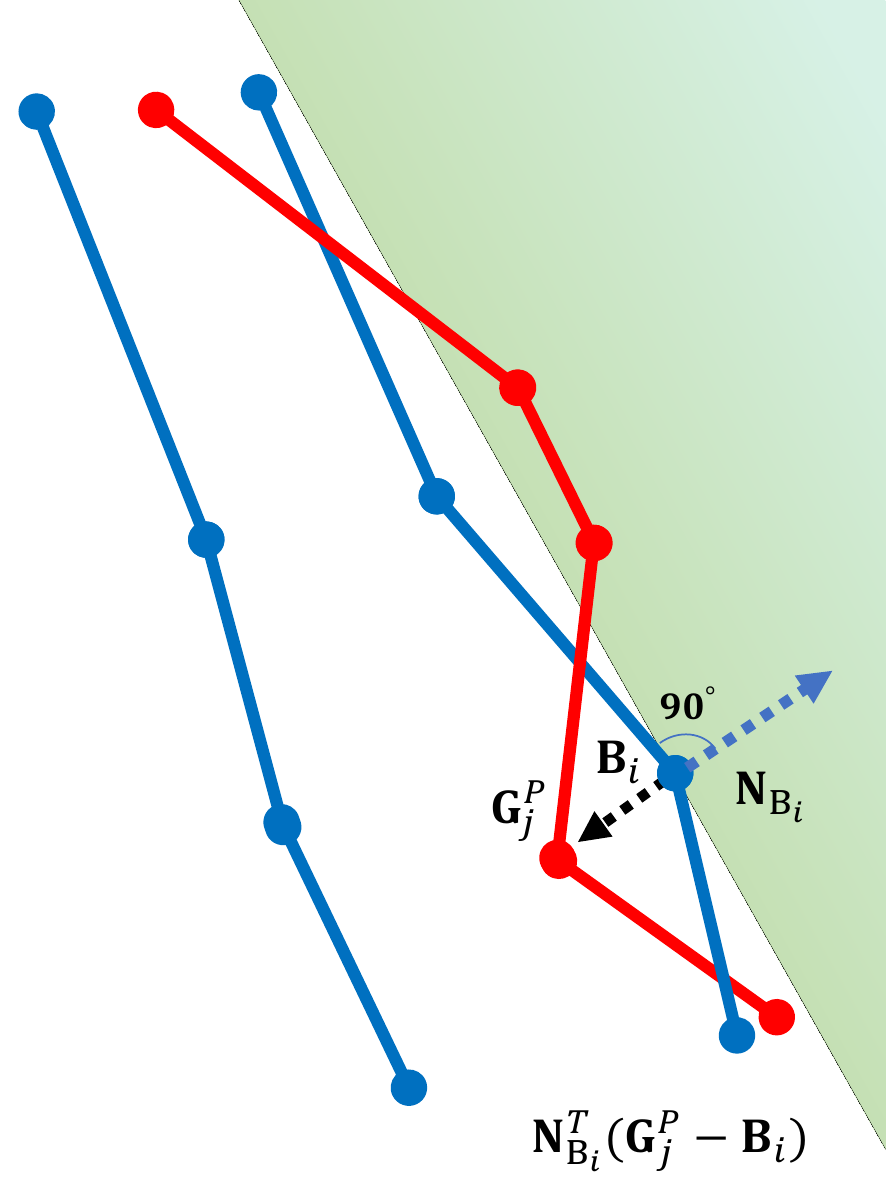}
		\caption{}
		\label{fig:penetAndBendA}
	\end{subfigure}
	\begin{subfigure}[b]{0.40\columnwidth}
		\includegraphics[width=\textwidth]{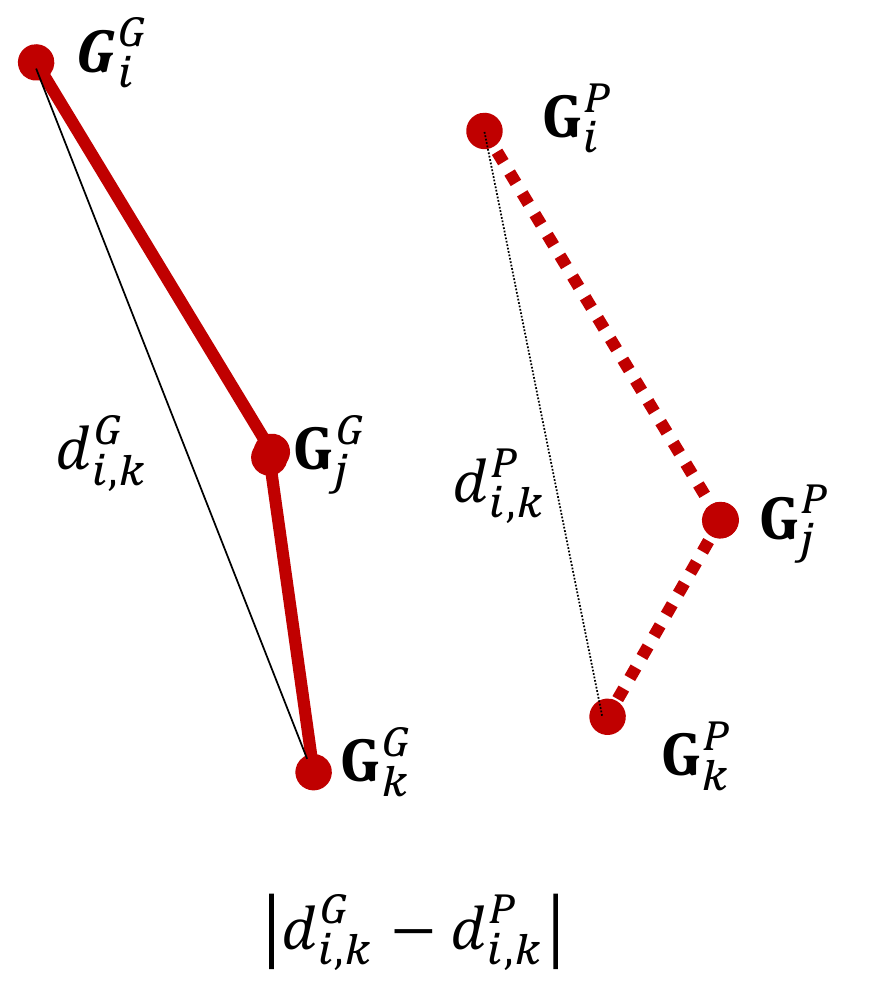}
		\caption{}
		\label{fig:penetAndBendB}
	\end{subfigure}
	\vspace{-3mm}
	\caption{\small {\bf Interpenetration and Bending loss terms.} \textbf{(a)} The interpenetration term $L_{pen}$ penalizes a garment vertex $\bG_j^P$ for being on the wrong side of the corresponding body point $\bB_i$. \textbf{(b)} The bending term $L_{bend}$ penalizes the distance between two neighbors of $\bG_j^P$ to differ from that in the ground truth.}
	\label{fig:penetAndBend}
\end{figure}

\parag{Interpenetration Term.}
To assess whether a garment vertex is inside the body, we first find the  nearest body vertex. At each iteration of the training process, we perform this search for all garment vertices. This yields $\mathcal{C}(\mB,\mG^P)$, a set of garment-body index pairs. \comment{one for the garment vertices and the other for the corresponding body vertices.}
We write $L_{pen}$ as
\begin{small}
\begin{align}
\!\!\!\!\!\!\!\sum_{\{i,j\}\in\mathcal{C}(\mB,\mG^P)}\!\!\!\!\!\!\!\!\!\!\! \bOne_{\{\|\bG_j^{P}-\bG_j^{G} \| <d_{tol}\}} \! ReLU(-\bN_{B_i}^T (\bG_j^{P}-\bB_i)) / N , 
	\label{eq:lossPen}
\end{align}
\end{small}
to penalize the presence of garment vertices inside the body. Here, $\bN_{B_i}$ is the normal vector at the $i^{th}$ body vertex, as depicted by Fig~\ref{fig:penetAndBendA}. This formulation penalizes garment vertex $G_j^P$ for not being on the green subspace of its corresponding body vertex $B_i$, provided that it is less than a distance $d_{tol}$ from its ground-truth position. In other words, the constraint only comes into play when the vertex is sufficiently close to its true position to avoid imposing spurious constraints at the beginning of the optimization. The loss term also penalizes traingle-triangle intesections between the body and the garment, which could happen when two neighboring garment vertices are close to the same body vertex. Unlike in~\cite{Guan12}, we do not force the garment vertex  to be within a predefined distance of the body because, in some cases, garment vertices can legitimately be far from it.

\parag{Normal Term.}
We write $L_{norm}$ as
\begin{equation}
	\label{eq:lossNormal}
	\frac{1}{N_F} \sum\limits_{i=1}^{N_F} {\left(1-{\left( \bF_i^{G}  \right) }^T \bF_i^{P}\right)^2},
\end{equation}
to penalize the angle difference between the ground-truth and predicted facet normals. Here, $N_F$, $\bF_i^{G}$ and $\bF_i^{P}$ are the number of facets, the normal vector of the $i^{th}$ ground-truth facet and of the corresponding predicted one, respectively.  

\parag{Bending Term.}
We take $L_{bend}$ to be
\begin{equation}
	\label{eq:lossBend}
	\frac{1}{\lvert \mathcal{N}_2 \rvert}\!\! \sum\limits_{\{i,k\}\in \mathcal{N}_{2}} { \mid \| \bG_i^{P}-\bG_k^{P} \rVert - \| \bG_i^{G}-\bG_k^{G} \rVert \mid},
\end{equation}
to emulate the bending constraint of NvCloth~\cite{Nvcloth}, the PBS method we use, which is an approximation of the one in \cite{Muller07}. Here, $\mathcal{N}_2$ denotes a set of pairs of vertices connected by a shortest path of two edges. This term helps preserve the distance between neighboring vertices of a given vertex, as  shown in Fig.~\ref{fig:penetAndBendB}.  Although it is theoretically possible to consider larger neighborhoods, the number of pairs would grow exponentially.



\subsection{Implementation Details}
\label{sec:implementation}

To apply the skinning method of~\cite{Kavan07}, we compute the skinning weight matrix $\mW$ using Blender~\cite{Blender} given the pose information of the garment mesh. The garment stream employs 6 residual blocks depicted in Fig.~\ref{residualBlock} following the common practice of ResNet ~\cite{He16}. In each block, we adopt the mesh convolution layer proposed in~\cite{Verma18}, which uses $1$-ring neighbors to learn patch-wise features at each convolution layer.  As the mesh convolution operators rely on trainable parameters to weigh the contribution of neighbors,  we always concatenate the input vertex 3D locations to their input vectors so that the network can learn topology-dependent convolutions. While using the exact PointNet architecture of \cite{Qi17a} in the body stream, we observed that all point-wise body features converged to the same feature vector, which seems to be due to ReLU saturation. To prevent this, we use leaky ReLUs with a slope of $0.1$ and add a skip connection from the output of the first Spatial Transformer Network (STN) to the input of the second MLP block. To use the body features in the garment stream as shown in Fig.~\ref{garmentBranch}, the $512$-dimensional global body features are repeated for each garment vertex. For the local body pooling depicted by Fig.~\ref{LocalPooling}, we downscale the 3D body points along with their point-wise features by a factor $10$. This is done by average pooling applied to the point-wise body features with a $16$ neighborhood size. For the local max-pooling of body features in Fig.~\ref{LocalPooling}, the number of neighbors is $15$. {To increase the effectiveness of the interpenetration term in Eq.~\eqref{eq:lossPen}, each matched body point $B_i$ is extended in the direction of its normal vector by $20$\% of average edge length of the mesh to ensure that penetrations are well-penalized,} and the tolerance parameter $d_{tol}$ is set to $0.05$ for both our dataset and that of~\cite{Wang18}. Additional details are given in the supplementary material. To train the network, we use the PyTorch~\cite{PyTorch} implementation of the Adam optimizer~\cite{Kingma14a} with a learning rate of $0.001$. In all the experiments reported in the following section, we empirically set the weights of Eq.~\ref{eq:lossTotal}, $\lambda_{normal}$, $\lambda_{pen}$ and $\lambda_{bend}$ to $0.3$, $1.0$ and $0.5$.


\comment{
The first uses a  inspired architecture to extract local and global information about the person. Given a 3D garment template, the second stream exploits the global information  to compute point-wise, patch-wise and global features. These features, along with the global body ones, are then fed to a fusion subnetwork to predict the shape of the fitted garment. To further model the skin-cloth interactions, we introduce an auxiliary stream that performs nearest neighbor pooling of local body features at each garment point and feeds the pooled features to the fusion subnetwork. This makes it possible to model how the body deforms the cloth  and results in increased prediction accuracy. 

Following the order invariant operations of PointNet \cite{Qi17a}, the Multilayer Perceptrons (MLP) are shared among all of the vertices of $\mathcal{M}$ and $\mathcal{B}$, and each global part consists of max-pooling layer over all the points to have a global description of the 3D shapes. For the early processing of garment and body features, we use two Spatial Transformer Networks in the way similar to the architecture in \cite{Qi17a}. For each mesh convolution layer, we use the convolution operation described in \cite{Verma18} with $1$-ring neighborhood. The details of the garment network is depicted in , where global body features (blue colored) are inputs of each MLP block by skip connections. On the other hand, the body branch of our network is a variant of PointNet \cite{Qi17a} with slight modifications.

\subsection{Local Pooling of Body Features}
\label{sec:locaPooling}

\begin{figure}[t]
\centering
	\centering
	\includegraphics[width=0.25\textwidth]{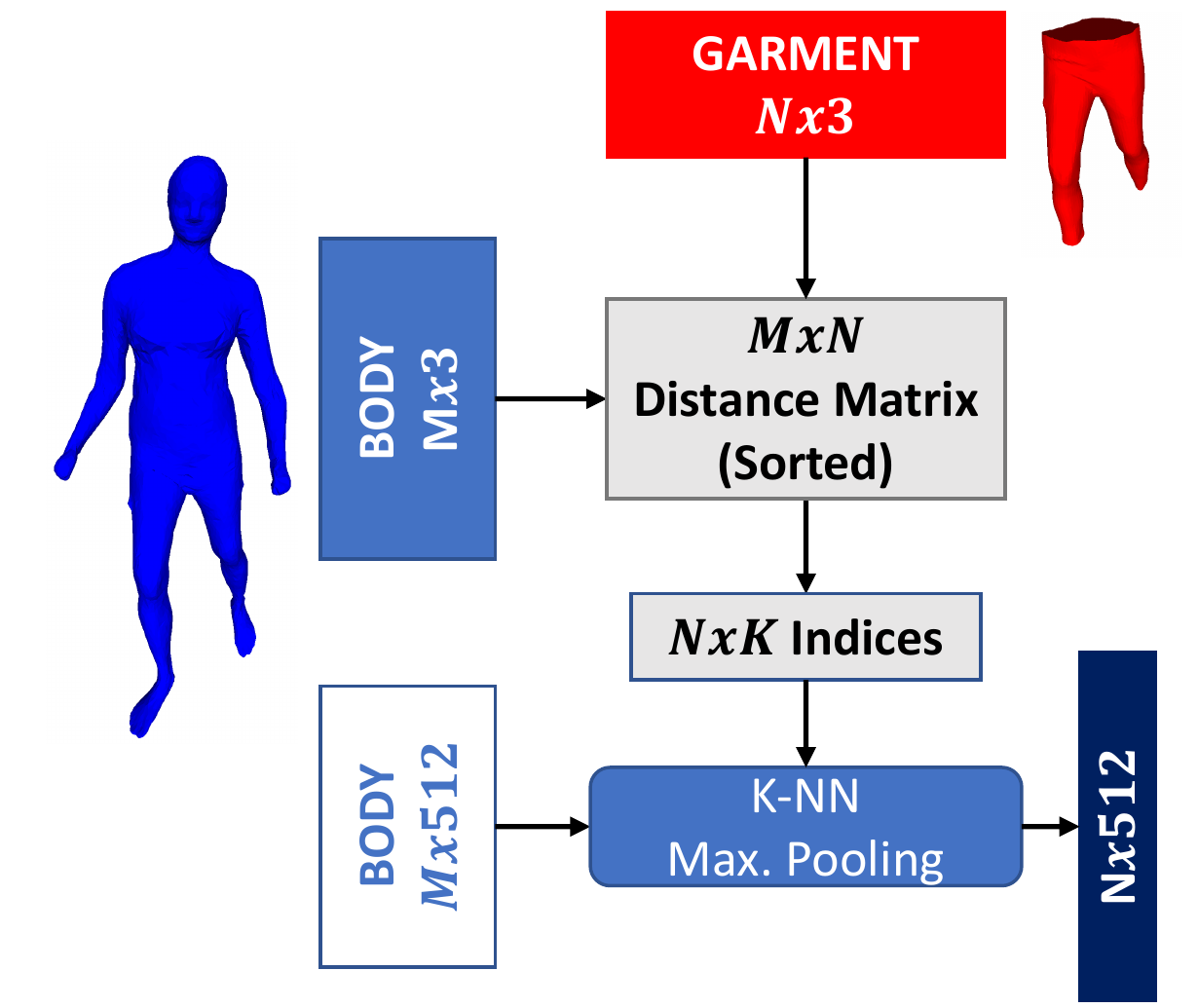}
	\vspace{-3mm}
	\caption{\small {\bf $K$ nearest neighbor pooling in Fig.~\ref{fig:streamsB}.} We compute the $K$ nearest neighbor body vertices of each garment vertex and max-pool their local features. 
	}
	\label{LocalPooling}
\end{figure}

The middle part in Figure \ref{fig:streamsB} finds the body vertex indices to be pooled by extracting the $K$-nearest body neighbors (NN) around each garment vertex . The warped input garment is approximately aligned with the target body. Therefore, one can select the body features by relying on the proximity of garment vertices to those of body.

A diagram of our pooling operation is depicted in Figure \ref{LocalPooling}. To reduce the computational complexity, as in the grouping and PointNet layers of PointNet++ \cite{Qi17b} the body features ($N\times512$) and 3-D body locations are first downscaled to $M$ points where $M<<N$. We first compute and sort the distance matrix $D \in \mathcal{R}^{M\times N}$ of the $M\times 3$ body and $N\times 3$ garment locations. Then, the index set $I_D \in \mathcal{R}^{N\times K}$  corresponding to the first $K$ rows of $D$ is constructed. By using $I_D$ and $M\times 512$ body features, we compute the body tensor $N\times K\times 512$ where the second dimension is the number of neighboring body vertices. This tensor is max-pooled over the second dimension, and given as input (dark blue in Figure \ref{fig:streams}) for the final processing of the garment features in the MLP layers of the fusion block. Our experiments show that this local body feature pooling step for each garment vertex significantly improves fitting accuracy.
}


\section{Experiments}
\label{sec:experiments}

In this section, we evaluate the performance of our framework both qualitatively and quantitatively. We first introduce the evaluation metrics we use, and conduct extensive experiments on our dataset to validate our architecture design. Then, we compare our method against the only state-of-the art method~\cite{Wang18} for which the training and testing data is publicly available. Finally, we perform an ablation study to demonstrate the impact of our loss terms.

\subsection{Evaluation Metrics}
\label{sec:metrics}

We introduce the following two quality measures:
\begin{align}
	\mE_{dist}   & = \frac{1}{N}\sum\limits_{i=1}^N{\lVert \bG_i^{G}-\bG_i^{P} \rVert}  \; , \label{eq:eucEval1} \\
	\mE_{norm} & = \frac{1}{N_F}\sum\limits_{i=1}^{N_F} \arccos\left( \frac{  { ( \bF_i^{G} ) }^T \bF_i^{P}}{\| \bF_i^{G} \| \| \bF_i^{P} \|} \right)\; . \label{eucEval2}
\end{align}
$\mE_{dist}$ is the average vertex-to-vertex distance between the predicted mesh and the ground-truth one, while  $\mE_{norm}$ is the average angular deviation of the predicted facet normals to the ground-truth ones. As discussed in~\cite{Brouet12}, the latter is important because the normals are key to the appearance of the rendered garment.

\subsection{Analysis on our Dataset}  
\label{sec:resOurs}

We created a large dataset featuring various poses and body shapes. We first explain how we built it and then test various aspects of our framework on it. 

\parag{Dataset Creation.}
We used the Nvidia physics-based simulator NvCloth~\cite{Nvcloth} to fit a T-shirt, a sweater and a pair of jeans represented by 3D triangulated meshes with 10k vertices on synthetic bodies generated by the SMPL body model~\cite{Loper15}, represented as meshes with 6890 vertices.
To incorporate a variety of poses, we animated the SMPL bodies using the yoga, dance and walking motions from the CMU mocap~\cite{CMUHMC} dataset. The training, validation and test sets consist of $500$, $20$ and $80$ bodies, respectively. The T-shirt, the sweater and the jeans have, on average, $40$, $23$ and $31$ poses, respectively. To guarantee repeatability for similar body shapes and poses, each simulation was performed by starting from the initial pose of the input garment. 

\comment{Moreover, we averaged the results over at least $40$ iterations \MS{This sounds strange.} to guarantee obtaining a quasi-static result, thus avoiding to produce different fitting results for very similar bodies.}


\parag{Quantitative Results.}
Recall from Section~\ref{sec:arch} that we implemented two variants of our network, \Global{} that relies solely on global body-features and \Local{} that also exploits local body-features by performing nearest neighbor pooling as shown in Fig.~\ref{LocalPooling}. As the third variant, we implemented a simplified version of \Global{} in which we removed the mesh convolution layers that produce patch-wise garment features. It therefore performs only point-wise operations (\emph{i.e.} $1\times 1$ conv.) and max-pooling layer, and we dub it \Late{}, which can also be interpreted as a two-stream PointNet~\cite{Qi17a} with extra skip connections. We also compare against the garment warped by dual quaternion skinning (DQS)~\cite{Kavan07}, which only depends on the body pose.

\begin{figure}[htbp]
	\centering
	\setlength{\tabcolsep}{0pt} 
    \renewcommand{\arraystretch}{1.0} 
	\resizebox{\linewidth}{!}{%
	\begin{tabular}{cc}
		\includegraphics[width=0.24\textwidth]{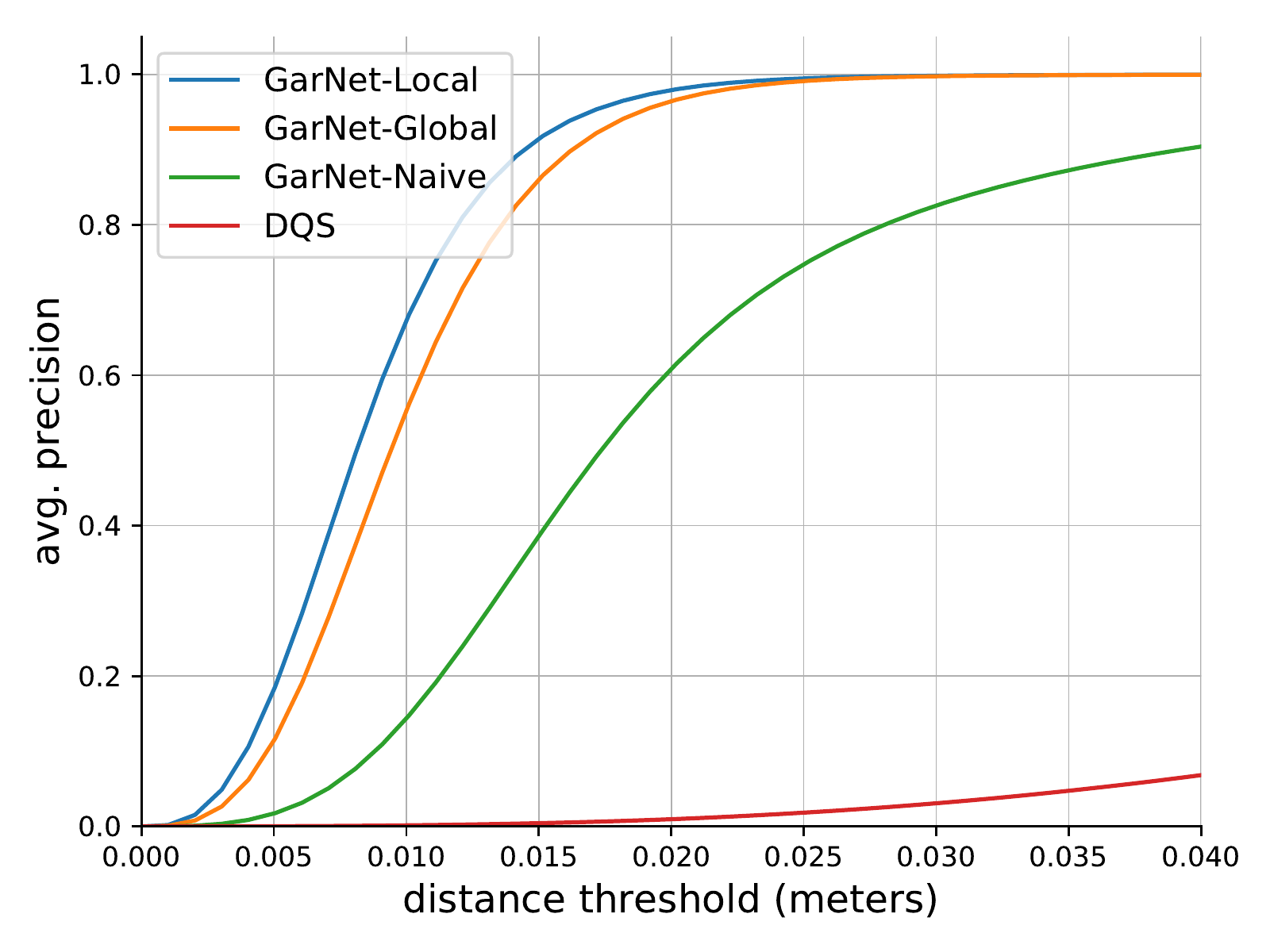}&
		\includegraphics[width=0.24\textwidth]{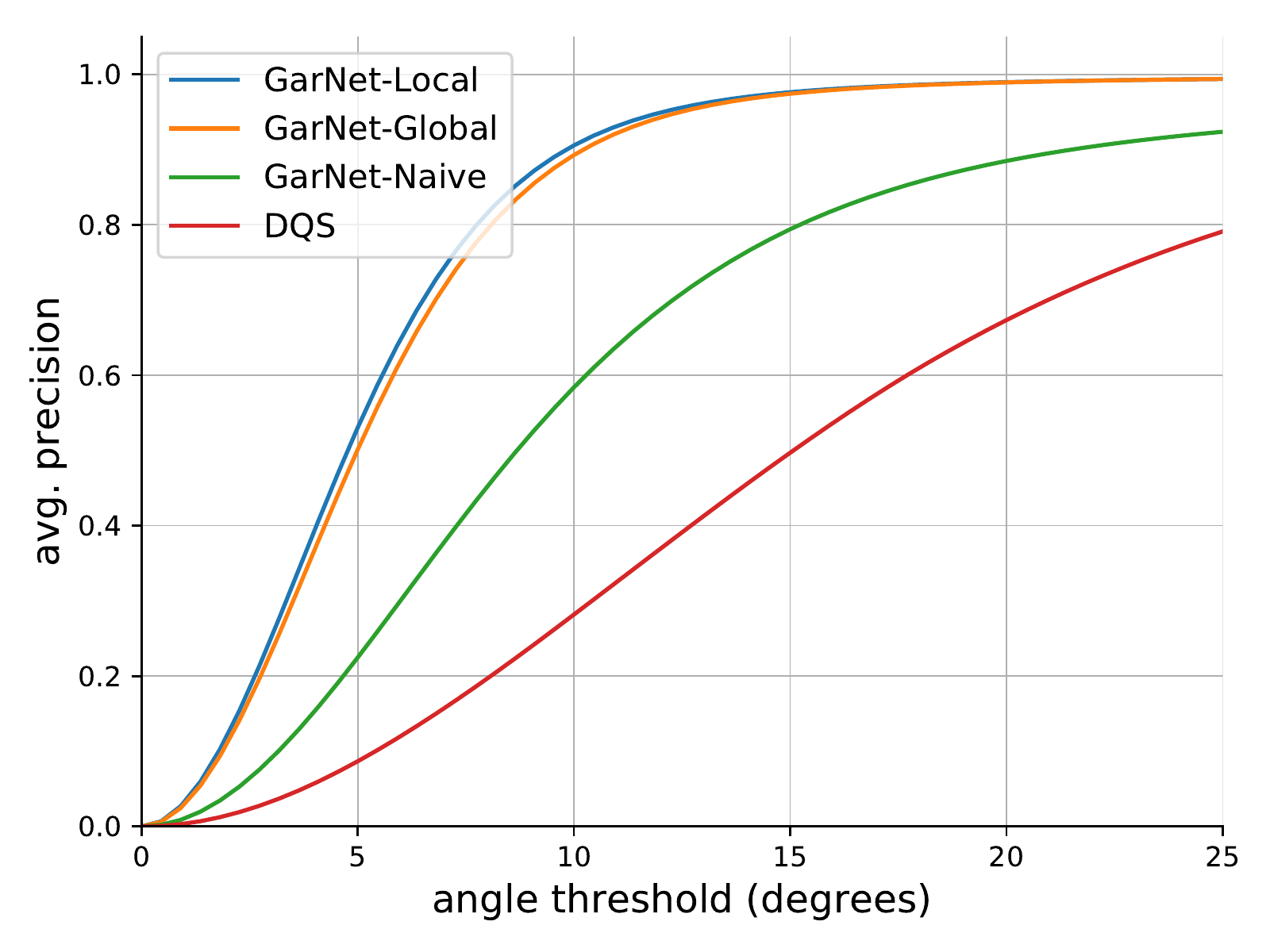}\\[-2mm]
		{\small Distance Jeans} & {\small Normals Jeans}\\
	\end{tabular}}
	\resizebox{\linewidth}{!}{%
		\begin{tabular}{cc}
		\includegraphics[width=0.24\textwidth]{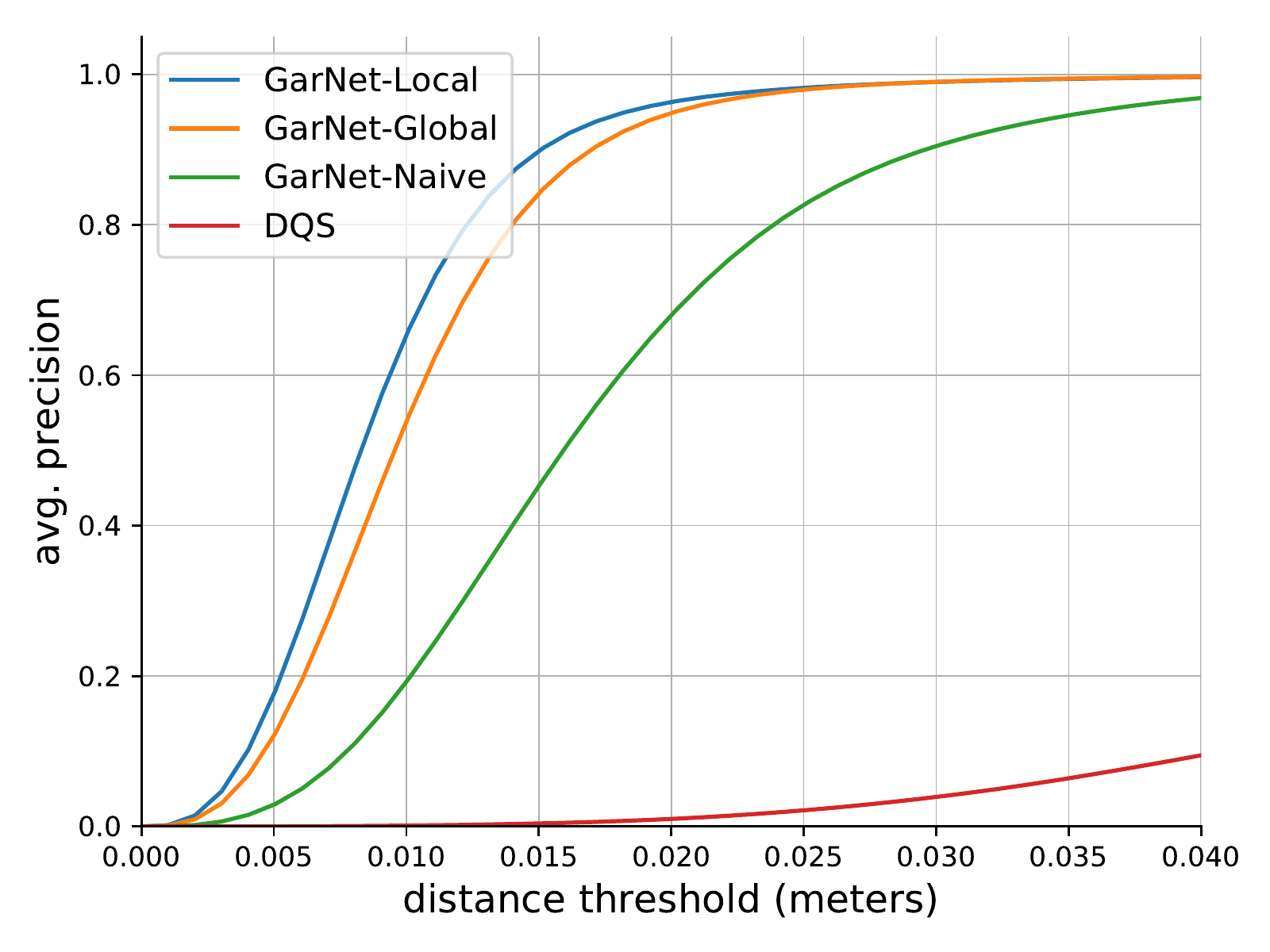}&
		\includegraphics[width=0.24\textwidth]{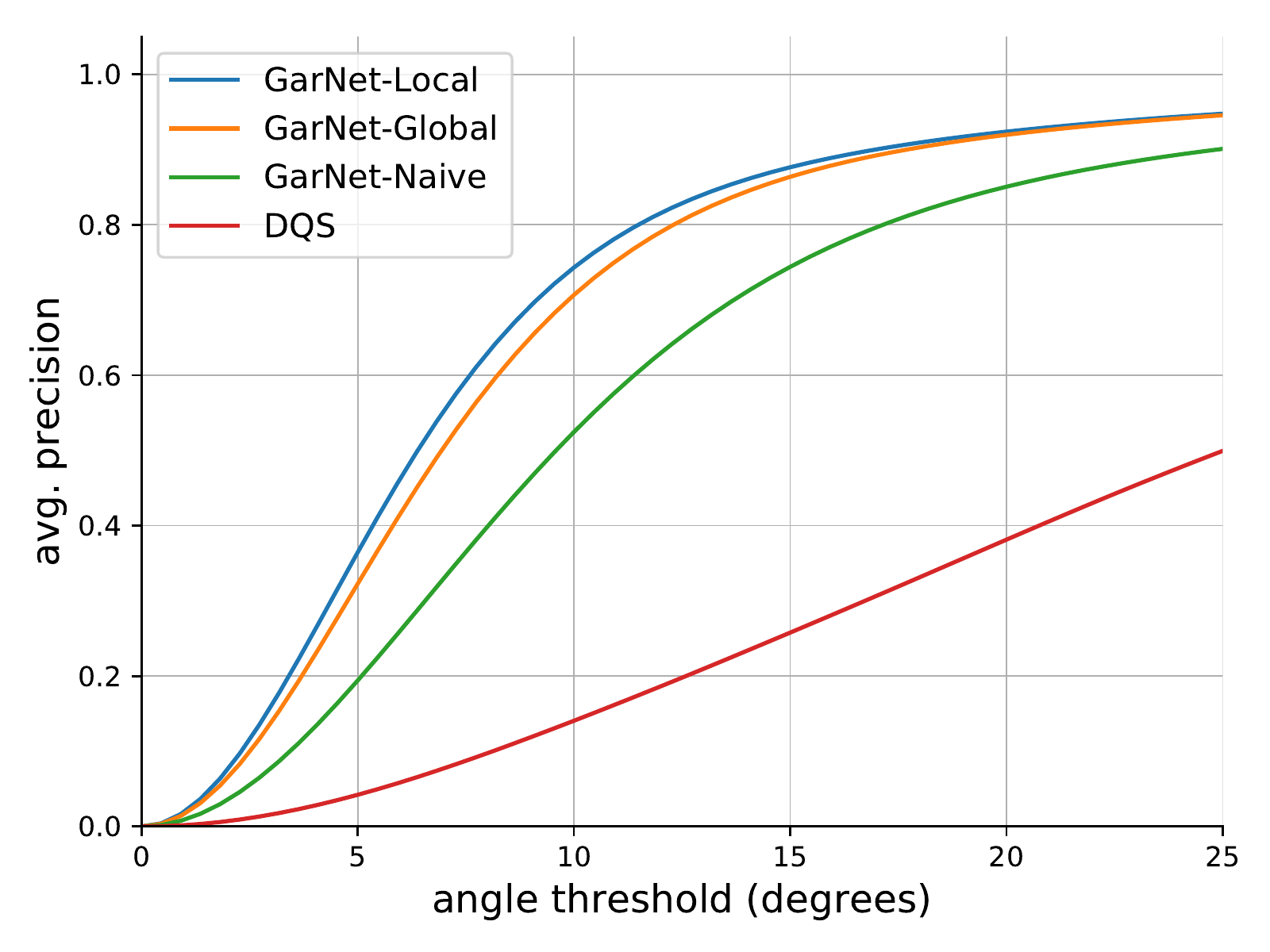}\\
		{\small Distance T-shirt}& {\small Normals T-shirt}\\
	\end{tabular}}
	\resizebox{\linewidth}{!}{%
		\begin{tabular}{cc}
		\includegraphics[width=0.24\textwidth]{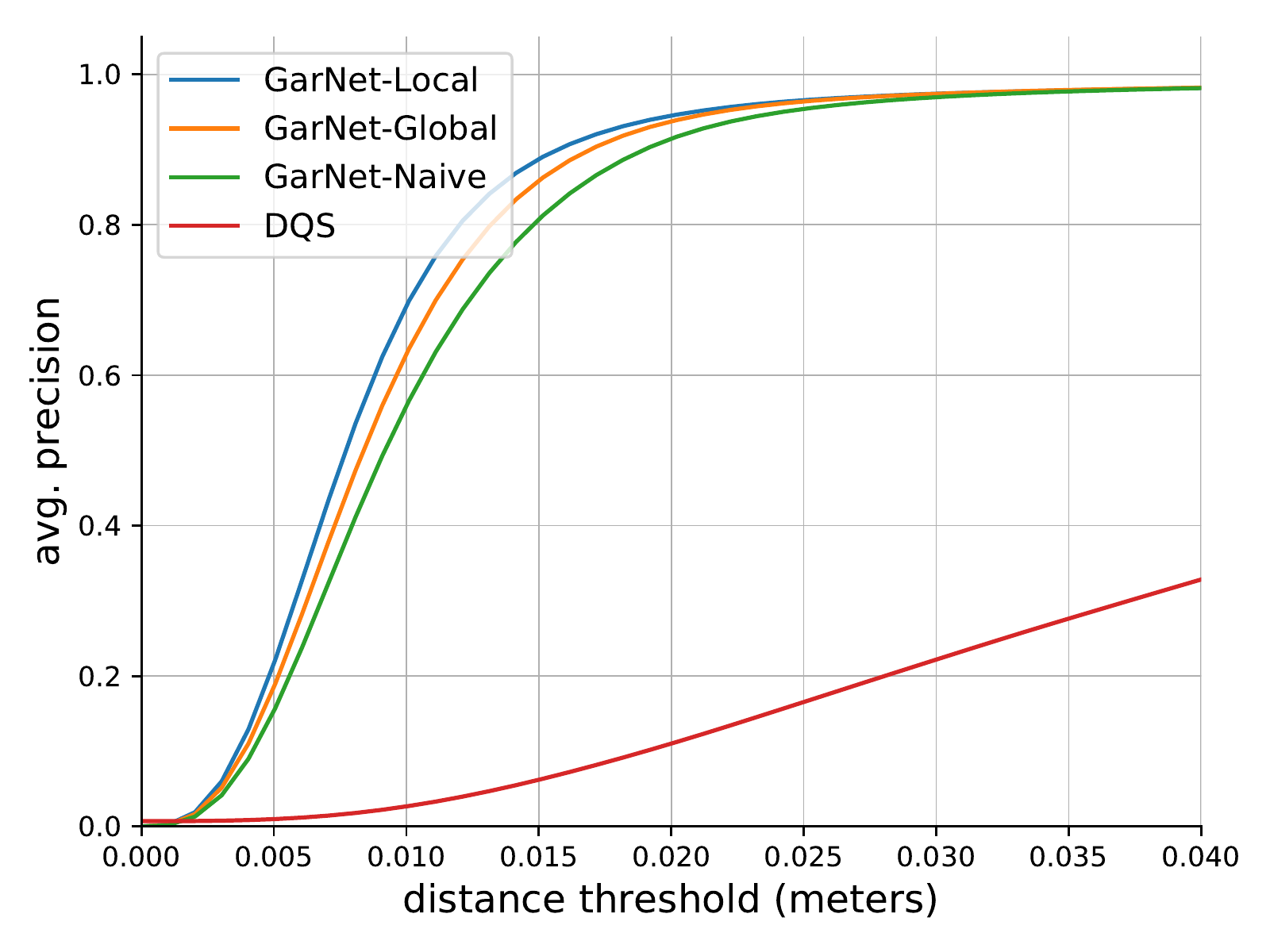}&
		\includegraphics[width=0.24\textwidth]{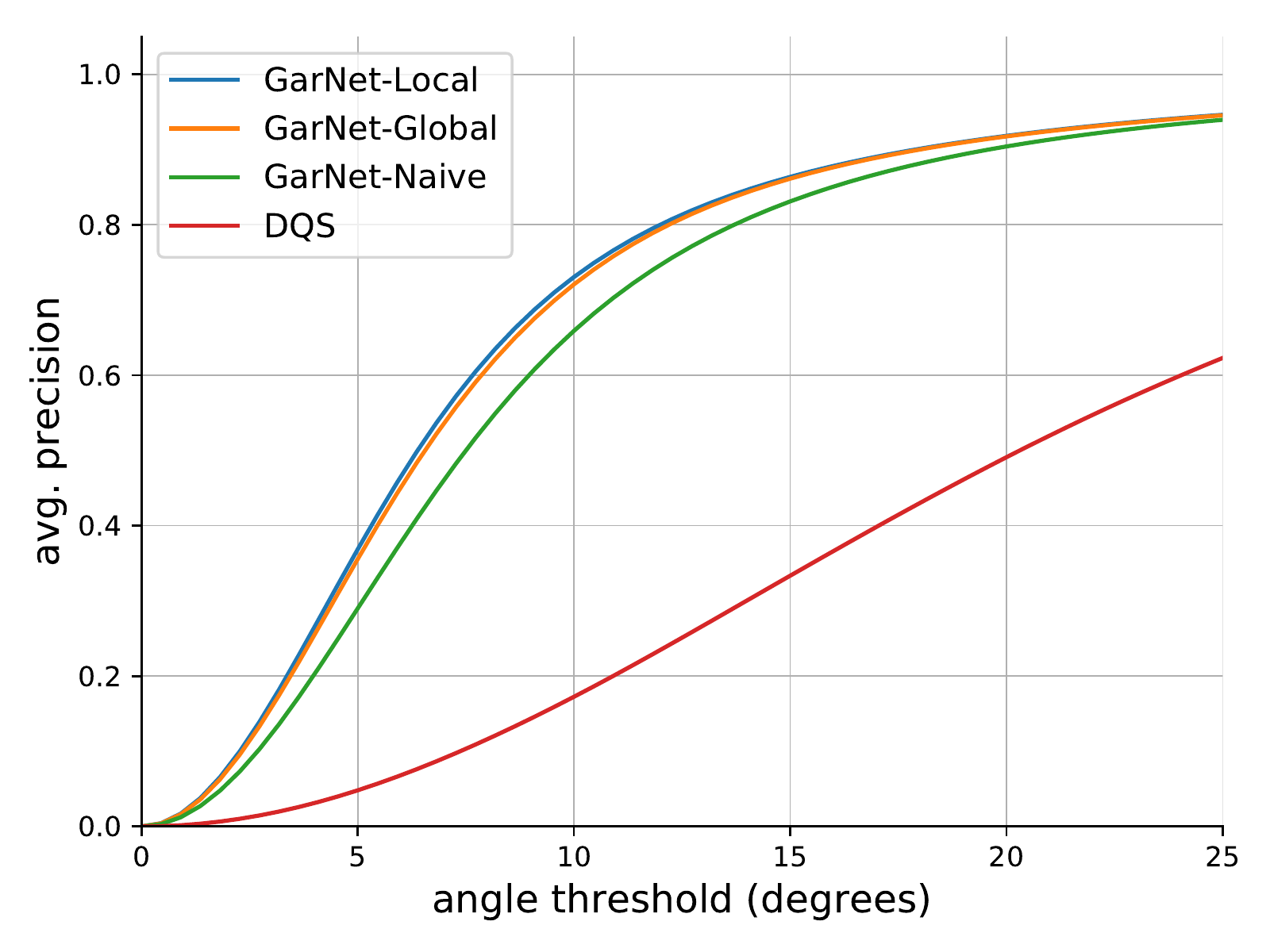}\\
		{\small Distance Sweater} & {\small Normals Sweater}\\
	\end{tabular}}

	\caption{Average precision curves for the vertex distance and the facet normal angle error.}
	\label{fig:precision}
\end{figure}

	

\begin{figure}[t!]
	\centering
		\includegraphics[width=0.95\linewidth]{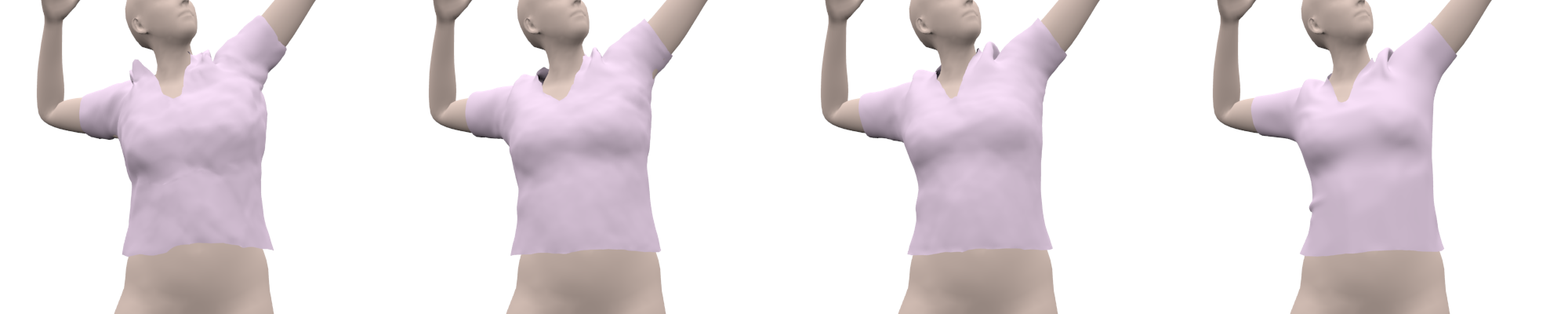}\\
		\includegraphics[width=0.95\linewidth]{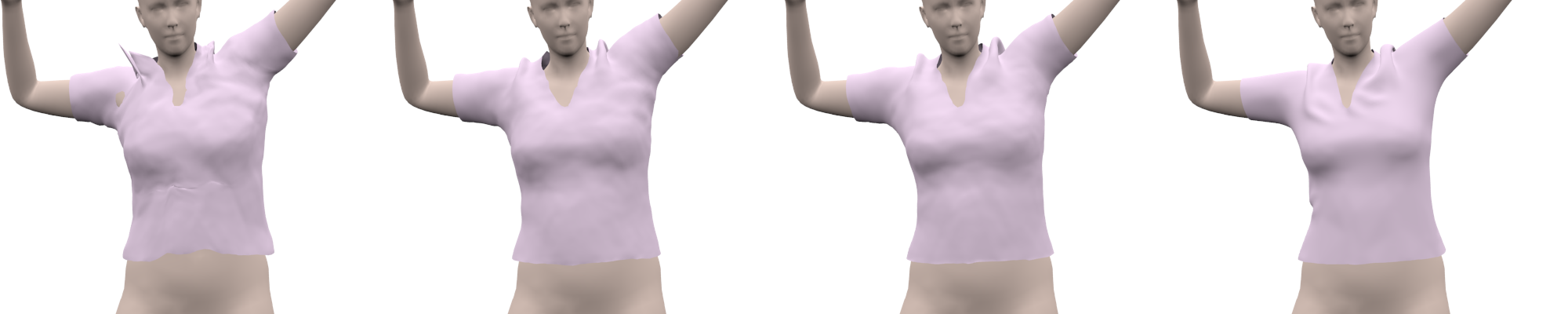}\\
	\begin{tabular}{cccc}
	\hspace{-7mm}{\small \Late{}}&\hspace{-2mm}{\small \Global{}}&\hspace{-2mm}{\small \Local{}}&\hspace{3mm}{\small \PBS{}}
	\end{tabular}
	\vspace{-3mm}
	\caption{\small {\bf Comparison on the T-shirt}. \Late{} produces artifacts near the shoulder while \Local{},~\Global{} and \PBS{} yield similar results.}
	\label{fig:archTshirts}
\end{figure}


\begin{table}[h!]
	\centering
	\resizebox{\columnwidth}{!}{%
\begin{tabular}{c|c|c|c}
            & \textbf{Jeans} & \textbf{T-shirt}  & \textbf{Sweater} \\ \hline
            & $\mathcal{E}_{dist}/\mathcal{E}_{norm}$  & $\mathcal{E}_{dist}/\mathcal{E}_{norm}$ & $\mathcal{E}_{dist}/\mathcal{E}_{norm}$ \\ \hline
\Local    & \textbf{0.88}/\textbf{5.63} & \textbf{0.93}/\textbf{8.97} & \textbf{0.97}/\textbf{9.21}  \\ \hline
\Global   & {1.01}/5.85 & 1.05/9.48 & {1.03}/9.36  \\ \hline
\Late    & 2.13/12.59 & 1.78/13.48 & 1.13/10.3  \\ \hline
DQS \cite{Kavan07}   & 11.43/22.0 & 9.98/30.74 & 6.47/24.64
\end{tabular}}
\vspace{-3mm}
\caption{Average distance in cm and face normal angle difference in degrees between the PBS and predicted vertices.}
\label{tableAll}
\end{table}

In Table \ref{tableAll}, we report our results in terms of the $\mE_{dist}$ and $\mE_{norm}$ of Section~\ref{sec:metrics}. In Fig.~\ref{fig:precision}, we plot the corresponding average precision curves for T-shirts, jeans and sweaters. The average precision is the percentage of vertices/normals of all test samples whose error is below a given threshold. \Late{} does worse than the two others, which underlines the importance of patch-wise garment features. \Global{} and \Local{} yield comparable results with an overall advantage to \Local{}. Finally, in Table~\ref{tableTiming}, we report the computation times of our networks and of the employed PBS software. Note that both variants of our approach yield a 100$\times$ speedup.
\begin{table}[]
\resizebox{\columnwidth}{!}{%
\begin{tabular}{c|c|c|c|c|c}
          & \Local & \Global & \Late & PBS & PBS$^\dagger$ \\ \hline
time (ms) &    68   &     59   &   0.2  &   $>$ 19000 & $>$7200
\end{tabular}}
	\caption{Comparison of the computation time. We used a single Nvidia TITAN X GPU for PBS and for our networks. In our case, forward propagation was done with a batch size of $16$. PBS$^\dagger$ stands for PBS computation excluding the time spent during the warping of template garment onto the target body pose.}
	\label{tableTiming}
\end{table}

\parag{Tests on unseen poses.}
The T-shirt dataset is split such that 50\% (25\%) of the poses (uniformly sampled within each motion) are in the training set; the rest are in the test set. The distance and angle errors increases to $1.16$ ($1.68$) cm and $9.71$ ($11.88$)$^\circ$. Since our poses are carefully sampled to ensure diversity, the performance on the splits above indicate generalization ability.

\parag{Qualitative Results.} 
Fig.~\ref{fig:archTshirts} depicts the results of the \Local, \Global~and \Late{} architectures. The \Global{} results are visually similar to the \Local{} ones on the printed page; however, \Global~produces a visible gap between the body and the garment while the garment draped by \Local~is more similar to the PBS one. \Late{} generates some clearly visible artifacts, such as spurious wrinkles near the right shoulder. By contrast, the predictions of \Local{} closely match those of the PBS method while being much faster. We provide further evidence of this in Fig.~\ref{fig:visRes} for three different garment types. Additonal visual results are provided in the supplementary material.

\begin{figure*}[htbp]
	\centering
	\begin{tabular}{c}
	\includegraphics[width=0.80\textwidth]{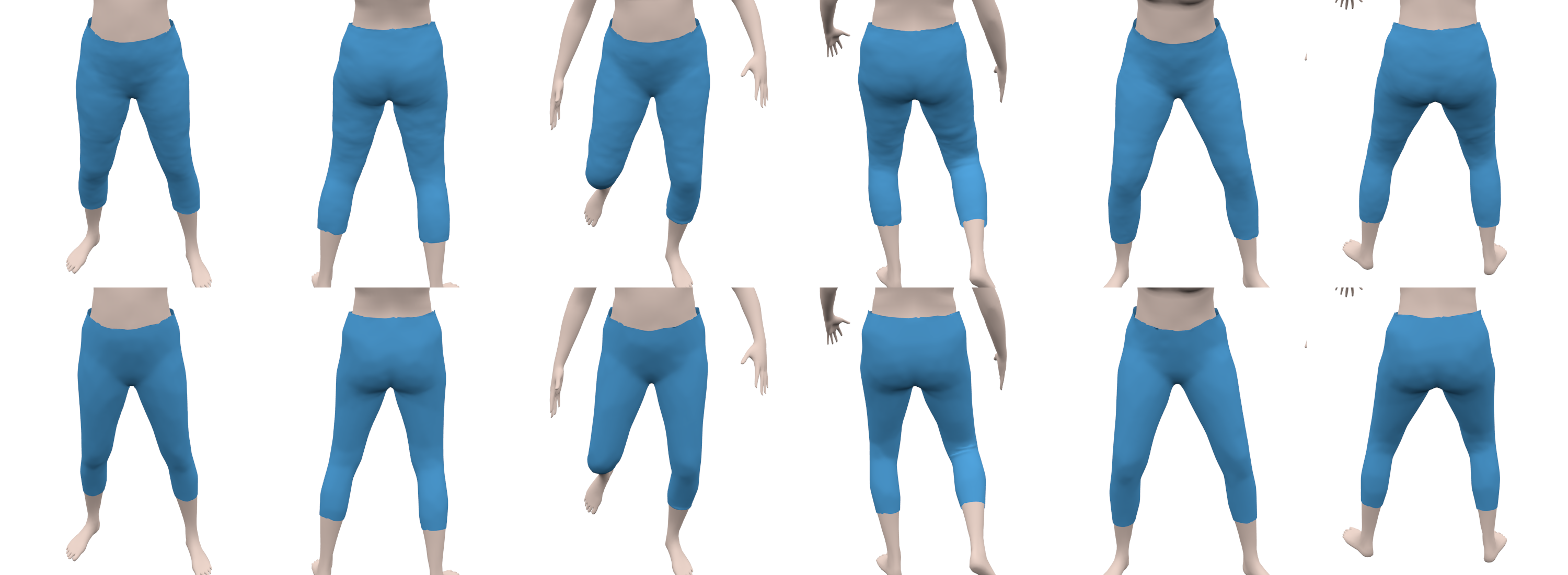}\\
	\includegraphics[width=0.80\textwidth]{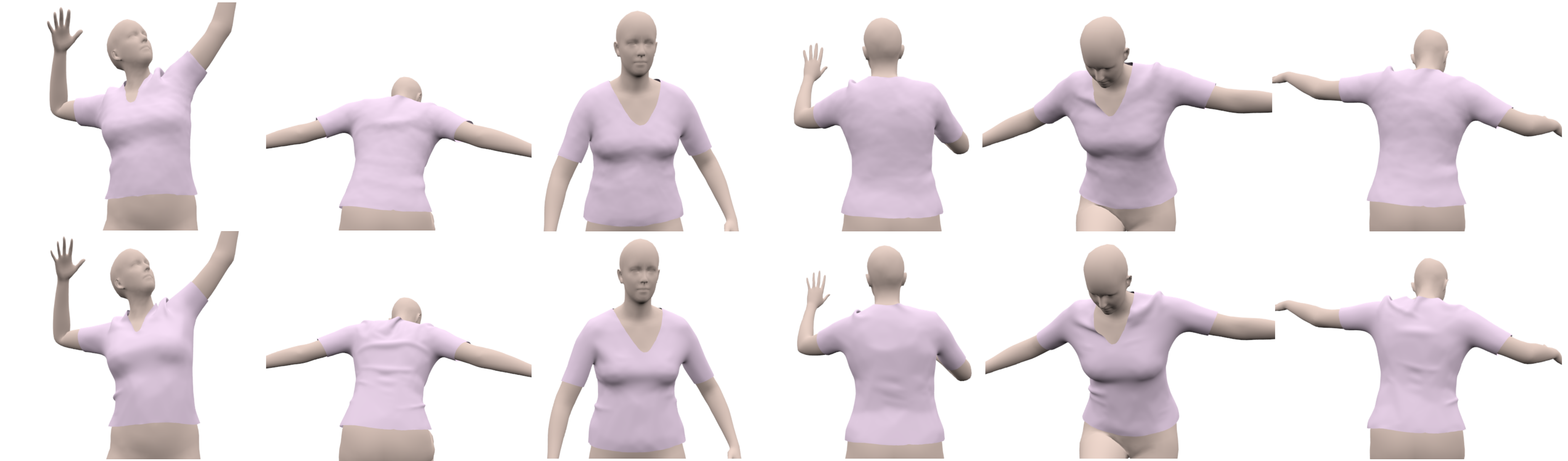}\\
	\includegraphics[width=0.80\textwidth]{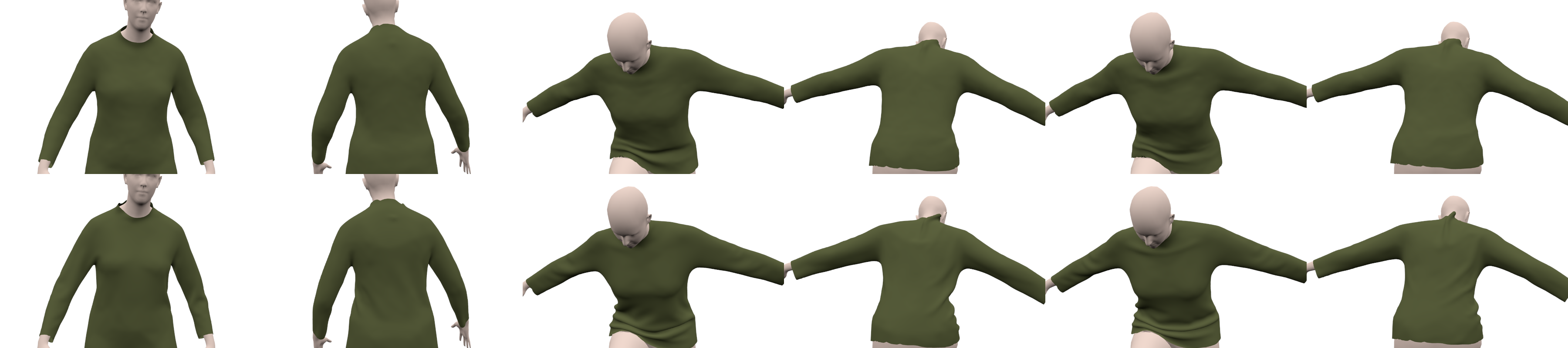}\\
	\end{tabular}
	\vspace{-3mm}
	\caption{\textbf{ \Local{} (top) vs \PBS{} (bottom) results for several poses.} Note how similar they are, even though the former were computed in approx. $70$ms instead of $20$s. Our method successfully predicts the overall shape and details with intermediate frequency.}
	\label{fig:visRes}
\end{figure*}

\subsection{Results on the Dataset of \cite{Wang18}}
\label{sec:resUCL}
As discussed in Section~\ref{sec:related},~\cite{Wang18} is the only non-PBS method that addresses a problem similar to ours and for which the data is publicly available. Specifically, the main focus of~\cite{Wang18} is to drape a garment on several body shapes for different garment sewing patterns. Their dataset contains 7000 samples consisting of a body shape in the T-pose, sewing parameters, and the fitted garment. Hence, the inputs to the network are the body shape and the garment sewing parameters. To use \Garnet{} for this purpose,  we take one of the fitted garments from the training set to be the template input to our network, and concatenate the sewing parameters to each vertex feature before feeding them to the MLP layers of our network. The modified architecture is described in more detail in the supplementary material.
We use the same training (95\%) and test (5\%) splits  as in~\cite{Wang18} and compare our results with theirs in terms of the normalized $L^2$ distance percentage, that is, $100\times {\lVert G^{G}-G^{P} \rVert}/{\lVert G^{G} \rVert}$, where  $G^{G}$ and $G^{P}$ are the vectorized ground-truth and predicted vertex locations normalized to the range $[0,1]$. We use this metric  here because it is the one reported in~\cite{Wang18}. As evidenced by Table \ref{tableUCL}, our framework generalizes to making use of garment parameters, such as sewing patterns, and significantly outperforms the state-of-the-art one of~\cite{Wang18}.
\begin{table}
\centering
\resizebox{0.7\columnwidth}{!}{%
\begin{tabular}{c|c|c|c}
& \Local & \Global & \cite {Wang18}\\ \hline
Dist. \%                & \textbf{0.89}    & 1.15    & 3.01  \\ \hline
Angle. $\sphericalangle$ & \textbf{7.40} & 7.53 & N/A
\end{tabular}}
	\caption{{\small Distance \% and angle error on the shirt dataset of \cite{Wang18}.}}
	\label{tableUCL}
\end{table}

\begin{figure}
	\centering
	\includegraphics[width=0.45\textwidth]{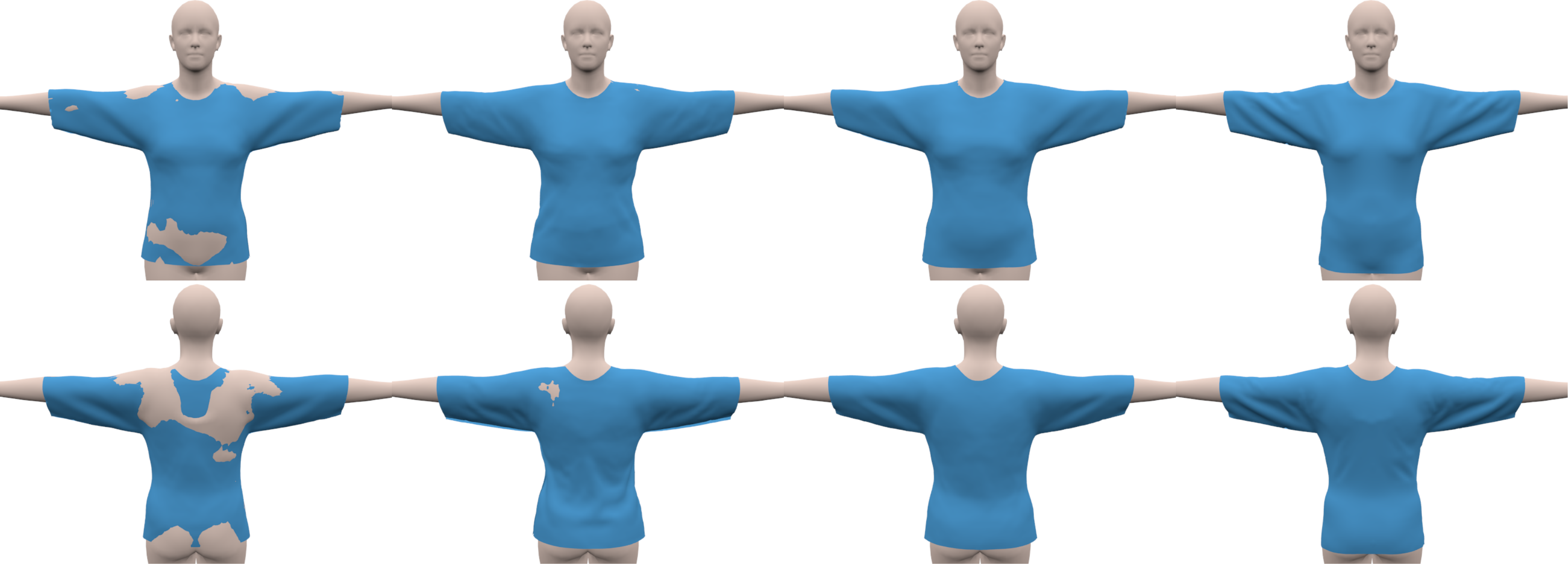}
	\begin{tabular}{cccc}
	\hspace{-0.3cm}{\small No penet.}&\hspace{0.3cm}{\small No norm.}&\hspace{0.3cm}{\small Full Loss}&\hspace{0.7cm}{\small PBS}
	\end{tabular}
	\vspace{-3mm}
	\caption{{\bf Ablation study}. Reconstruction without some of the loss terms results in interpenetration (left) or different wrinkles at the back (second from  left). By contrast, using the full loss yields a result very similar to the PBS one (two images on the right).}
	\label{AblationVisual}
\end{figure}

\parag{Ablation study.}
We conducted an ablation study on the dataset of~\cite{Wang18} to highlight the influence of the different terms in our loss function. We trained the network by individually removing the penetration, bending, and normal term. We also report results without both the normal and bending terms. As shown in Table~\ref{tableAblation}, using the normal and bending terms significantly improves the angle accuracy. This is depicted in Fig.~\ref{AblationVisual} where the normal term helps remove the spurious wrinkles. While turning off the penetration term has limited impact on the quantitative results, it causes more severe interpenetration, as shown in Fig.~\ref{AblationVisual}.

\begin{table}[t!]
	\begin{tabular}{c|c|c}
		\textbf{Loss Function} & $\mathcal{E}_{dist}$ & $\mathcal{E}_{normal}$ \\ \hline
		$L_{vertex}+L_{pen}$ & \textbf{0.55} & 8.88 \\ \hline
		$L_{vertex}+L_{pen}+L_{bend}$      & 0.67 & 9.90 \\ \hline
		$L_{vertex}+L_{norm}+L_{bend}$      & 0.69 & 7.39\\ \hline
		$L_{vertex}+L_{pen}+L_{norm}$     & 1.08 & 7.40\\ \hline
		$L_{vertex}+L_{pen}+L_{norm}+L_{bend}$    & 0.72 & \textbf{7.36}
	\end{tabular}
	\caption{Ablation study on the shirt dataset of~\cite{Wang18}.}
	\label{tableAblation}
\end{table}



\section{Conclusion}

In this work, we have introduced a new two-stream network architecture that can drape a 3D garment shape on different target bodies in many different poses, while running 100 times faster than a physics-based simulator. Its key elements are an approach to jointly exploiting body and garment features and a loss function that promotes the satisfaction of physical constraints. By also taking as input different garment sewing patterns, our method generalizes to accurately draping different styles of garments.

Our model can drape the garment shapes to within 1 cm average distance from those of a PBS method while limiting interpenetrations and other artifacts. However, it still has a tendency to remove high-frequency details, as also observed in~\cite{Guan12,Santesteban19}, because regression tends to smooth. In future work, we will explore conditional Generative Adversarial Networks~\cite{Isola17} to add subtle wrinkles to further increase the realism of our reconstructions, as in \cite{Lahner18}. Another avenue of research we intend to investigate is mesoscopic-scale augmentation, as was done in~\cite{Beeler10}, to enhance the reconstructed faces.

\clearpage
{\small
\bibliographystyle{ieee_fullname}
\bibliography{string,vision,graphics,learning,optim,misc}
}

\end{document}